# A multi-task convolutional neural network for mega-city analysis using very high resolution satellite imagery and geospatial data

Fan Zhang[1], Bo Du[2] and Liangpei Zhang[1]

1. The State Key Laboratory of Information Engineering in Surveying, Mapping, and Remote Sensing, Wuhan University, P. R. China.

2. School of Computer, Wuhan University, P. R. China.

*Abstract*—**Mega-city analysis with very high resolution (VHR) satellite images has been drawing increasing interest in the fields of city planning and social investigation. It is known that accurate land-use, urban density, and population distribution information is the key to mega-city monitoring and environmental studies. Therefore, how to generate land-use, urban density, and population distribution maps at a fine scale using VHR satellite images has become a hot topic. Previous studies have focused solely on individual tasks with elaborate hand-crafted features and have ignored the relationship between different tasks. In this study, we aim to propose a universal framework which can: 1) automatically learn the internal feature representation from the raw image data; and 2) simultaneously produce fine-scale land-use, urban density, and population distribution maps. For the first target, a deep convolutional neural network (CNN) is applied to learn the hierarchical feature representation from the raw image data. For the second target, a novel CNN-based universal framework is proposed to process the VHR satellite images and generate the land-use, urban density, and population distribution maps. To the best of our** **knowledge, this is the first CNN-based mega-city analysis method which can process a VHR remote sensing image with such a large data volume. A VHR satellite image (1.2 m spatial resolution) of the center of Wuhan covering an area of 2606 km$^2$ was used to evaluate the proposed method. The experimental results confirm that the proposed method can achieve a promising accuracy for land-use, urban density, and population distribution maps.**

## I. INTRODUCTION

With the rapid economic growth and urbanization of China over the past three decades, many cities, such as Beijing, Shanghai, and Wuhan, have evolved into mega-cities which hold more than 10 million inhabitants. However, the dramatic growth of mega-cities often leads to severe urban problems (e.g., water supply, traffic congestion, air pollution, sewage disposal, and housing) because of the unplanned development and the growth of the urban population, especially in developing countries [1]-[4].

While the massive dynamics of urbanization often affect the ability to manage, organize, and build new settlements, it is an even impossible task to measure and record what has already happened in a mega-city. In order to handle these severe urban problems and achieve sustainable urban growth, government and city planners need accurate and efficient methods for the large-scale monitoring of city area at a fine scale. Consequently, it is important to understand and analyze the land-use cover of the city, delineate the areas of human settlement, and estimate the urban density or population density to effectively estimate the budgets and improve the condition of urban life [5].

There have been many attempts over the past years to understand and analyze the urban land-use classification, urban density, and population density estimation tasks [6]-[9]. The traditional approach to mapping urban land use or urban density is to manually delineate remote sensing images and use building footprint map to produce maps. However, the initial costs of collecting land-use information and building footprint map and converting them into digital data, as well as the ongoing costs of updating land-use or building footprint information, are difficult. Since acquiring these data is so costly, can we find other sources of urban area data?

Thanks to the recent advances in satellite sensors, remote sensing data have now become an option, and they can provide area-wide spatial data with the necessary spatial resolution to identify the characteristics of urban land use [10], urban density [11], and even population density [12]. The use of remote sensing data acquired by spaceborne or airborne sensors provide a high degree of practicability, automation, and transferability. A wide variety of algorithms have been proposed to effectively and efficiently monitor and analyze mega-cities (e.g., urban land-use classification, urban density, or population density estimation) from remote sensing images, especially very high resolution (VHR) images [7]-[9],[13].

For the urban land-use classification task, most of the existing approaches consist of two stages: a feature extraction (calculation) stage and a classification stage. A typical example is the bag-of-words (BoW) based method for scene classification [14] [15], which first extracts the low-level features (e.g., scale-invariant feature transform, local binary pattern, color histogram,) in an image, then counts the occurrences of the low-level features in the image, and finally trains a classifier to assign a land use type to the image according to its content. Some studies have also applied region-based approaches which are mainly based on blocks or districts separated by the main roads to extract the feature representation of physical land-cover structures [16]. The region-based strategies can extract different features (e.g., volume, footprint, vegetation fraction) in each block, generate a new feature representation for each block, and then train a classifier to assign a land use type to the block according to its features. Although these two-stage based approaches are highly efficient, they all require an elaborate manually designed framework and lots of expert knowledge to understand the relationship between the features and the land-use types. A large semantic gap exists between the manually designed low-level features and the high-level semantic meanings [17], which severely constrains the ability to tackle complex image scenes.

In addition to the land-use classification task, building and population densities are also important indices that can reflect the activities in urban areas. Both optical images and synthetic aperture radar (SAR) have been widely used to monitor urban areas. The traditional methods for estimating building or population density also consist of two main stages: feature selection or extraction and a manually designed formula for the building or population density estimation [18] [19]. Susaki et al. [8] proposed to use polarimetric SAR (PolSAR) images to estimate urban density. Their method is based on the method of Kajimoto and Susaki [20], which first extracts the homogeneous polarization orientation angle (POA) city districts and then normalizes the scattering-power components in each POA space to construct the final urban density. For population density estimation, Azar et al. [9] proposed to use both multi-resolution satellite imagery and geospatial data to generate a fine-scale population estimation. This method extracts the built-up area fraction and a population likelihood raster and is then combined with a manually designed formula for the fine-scale population distribution estimation. Similar to the land-use classification task, the building and population density task also needs a manually designed framework and lots of expert knowledge to understand the relationship between these features and the building/population density. A formula is also required to estimate the building or population density [8] [9], which hinders the adaptability to different study areas and data sources.

How to establish an efficient framework which can automatically extract the internal feature representation from the raw data and close the semantic gap between the remote sensing image and the high-level semantic meanings is still a critical problem. What is more, how to establish the relationship between multiple mega-city analysis tasks (land-use classification, urban and population density estimation) and integrate these tasks into a universal framework is an even more difficult challenge.

Artificial intelligence (AI) has recently become a thriving field with many practical applications and active research topics. Among them, deep learning methods, which allow computers to learn from the experience and understand the world in terms of a hierarchy of concepts, have drawn great attention and achieved great success in many different fields [21] [22]. Since the early 2000s, as one of the most prevalent deep learning methods, CNNs have been applied with great success to image classification, depth estimation, and recognition of objects [23]-[25]. A CNN has even been applied to a dataset of about a million images from the web that contained 1,000 different classes, achieving dramatic results [25].

Lots of attempts have been made to explore CNN-based methods for remote sensing applications. Zhang et al. [26] proposed the gradient boosting random convolutional network framework for large-scale remote sensing image scene classification. Yuan [27] proposed automatic building extraction in aerial scenes using a convolutional network, which makes it possible to better analyze the remote sensing images. Inspired by these successes, we aim to build a CNN-based remote sensing data analysis system to generate different-level products for mega cities, such as land-use maps, urban density estimation, and population density estimation, thereby providing a better understanding for city planning and development.

In this study, we first introduce a CNN model which can automatically extract the internal feature representation from the massive remote sensing data for mega-city analysis. We then propose a novel CNN-based universal framework which integrates the land-use map, urban density estimation, and population density estimation tasks into a single deep neural network, and simultaneously generates different-level products for the mega-city analysis. Finally, combined with the development of the GPU acceleration technique, a deep learning based application for mega-city analysis is presented which can generate a fine-scale analysis from the VHR remote sensing image and is able to process a very large data volume.

The rest of this paper is organized as follows. Section II outlines the study area and data sources. Section III presents the novel CNN-based universal framework for the land-use classification, urban density, and population density estimation tasks. The experiments are detailed in Section IV, along with the verification of the generated land use-map, urban density estimation, and population density estimation. The effectiveness of the proposed method is discussed in Section V, and the conclusions are drawn in Section VI.

## II. DATA

### A. *Study area*

Our research focused on the city of Wuhan in China (Fig. 1). Wuhan is located in central China (113°41'E–115°05'E，29°58'N–31°22'N) and covered an area of 8494 km$^2$. It is the capital of Hubei province and the largest city in central China with more than 10 million inhabitants. Wuhan is positioned at the confluence of the Yangtze and Han rivers, and is delineated by the rivers into three towns, i.e., Wuchang, Hankou, and Hanyang. Wuhan is recognized as the political, economic, financial, cultural, educational, and transportation center of central China. Influenced by several historical events (e.g., the Wuchang uprising, destruction during World War II, and the “Reform and Opening-Up” policy), various structure types and a complex land-use pattern have evolved. As a result of the huge city scale and the rapid development speed, an efficient and fast method for large-scale monitoring of the urban area is required for better city planning and sustainable urban growth. Therefore, we selected the central area of Wuhan, covering an area of 2606 km$^2$, as our study area, which is well suited for the goals of this research.

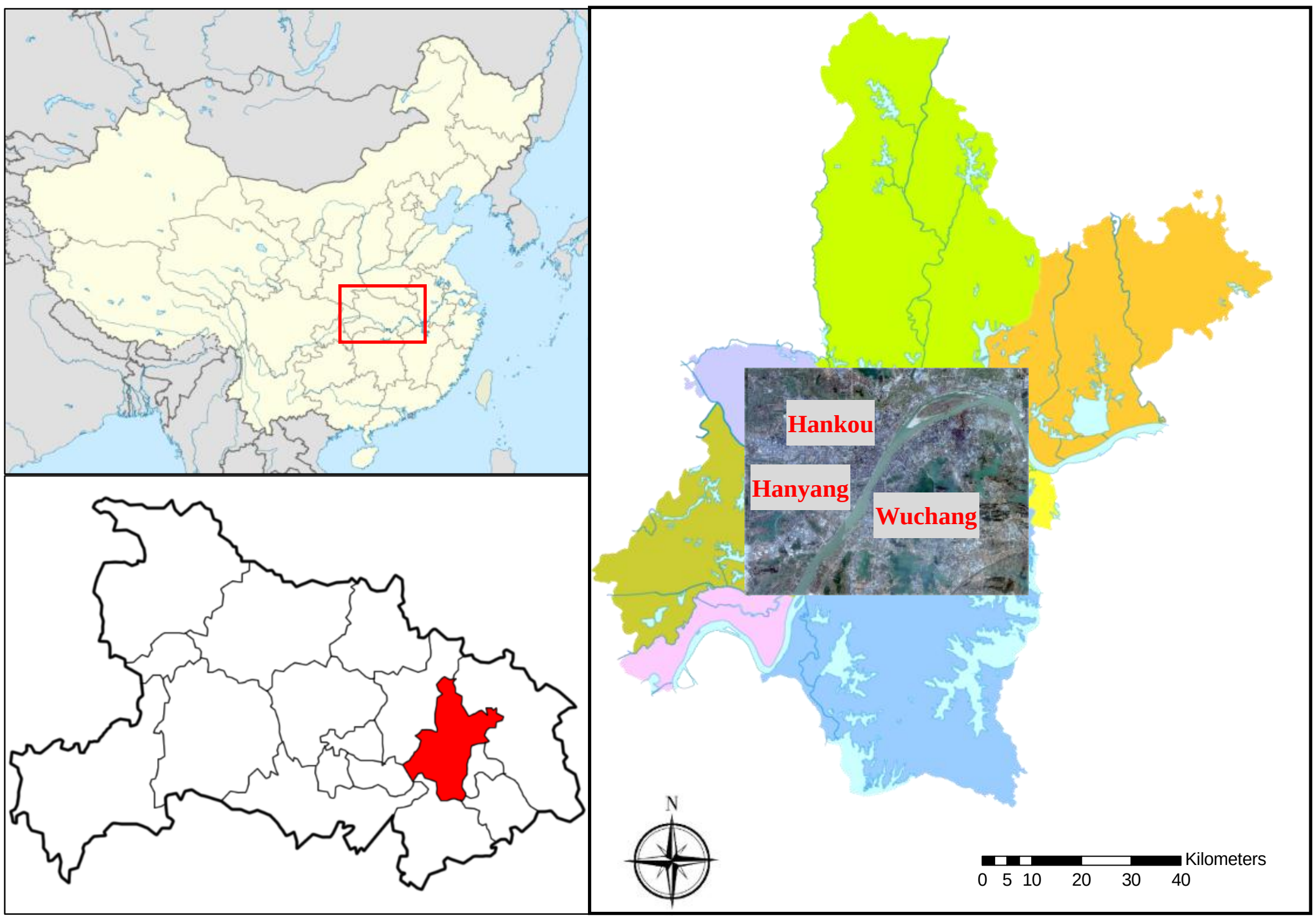

Fig. 1. The study area. Upper left: the location of Hubei province in China. Lower left: the administrative boundary of Wuhan in Hubei province. Right: remote sensing data from Wuhan acquired by DigitalGlobe in January 2016 (2606 km$^2$)

*B. Land-use data*

The original land-use data were obtained from the School of Urban Design, Wuhan University. These data were derived by the Wuhan Land Resources and Planning Bureau in 2014 according to the "Code for Classification of Urban Land Use and Planning Standards of Development Land" (Table I and Table

II) [28].

TABLE I

URBAN DEVELOPMENT LAND-USE CLASSES

| Primary types | | Secondary types | |
| --- | --- | --- | --- |
| Code | Name | Code | Name |
| R | Residential | R1 | Residential one |
| | | R2 | Residential two |
| | | R3 | Residential three |
| A | Administration and public services | | |
| B | Commercial and business facilities | | |
| M | Industrial | | |
| W | Logistics and warehouse | | |
| S | Road, street and transportation | | |
| U | Municipal utilities | | |
| G | Green space and square | | |

TABLE II

TOWN AND COUNTRY LAND-USE CLASS

| Primary types | | Secondary types | |
| --- | --- | --- | --- |
| H | Development land | H1 | Town and country development land |
| | | H2 | Regional transport facilities |
| | | H3 | Regional public services |
| | | H4 | Special use |
| E | Non-development land | E1 | Water area |
| | | E2 | Agriculture |
| | | E3 | Other |

Based on the “Code for Classification of Urban Land Use and Planning Standards of Development Land” and the land use data of the city of Wuhan, we designed a class hierarchy of 13 land-use classes, which can be defined to 10 main classes. The designed land-use classes are shown in Fig. 2.

The residential class is subdivided into three classes depending on the dominant building type, which can help us to further analyze the living conditions of Wuhan and estimate the population density. The “Residential one” class mainly includes areas of villas, and “Residential two” refers to modern residential areas which have a better living environment than “Residential three”. “Residential three”, which has become a severe problem in city planning, is a special type of Residential coming from the rapid and complicated socioeconomic development of China which is known as “villages in the city” [29].

The “Administration and public services” class mainly consists of administration buildings and the university campus area. The “Industrial” class areas are shaped by different building structures, ranging from huge manufactory to small buildings within one district, and feature little or no vegetation. For the transportation facilities, we defined two different types of land-use class. The “Road, street and transportation” class mainly refers to the major roads and overpasses. The “Regional transport facilities” class mainly refers to railways and railway stations. Additionally, we also defined two subclasses relating to water bodies and one main class relating to agricultural areas. Finally, due to the land-use complexity of Wuhan, we also defined a class containing the other land-use classes, such as bare soil or construction areas, to complete the designed land-use types.

The level of the designed land-use class hierarchy is mainly based on the “Code for Classification of Urban Land Use and Planning Standards of Development Land” and covers the requirements of both a holistic description of the city’s land use and also being transferable. In this study, we used a regular grid with a size of $200 \times 200$ pixels ($240$ m $\times 240$ m) to denote a land-use type, and the spatial resolution of each land-use type was 1.2 $m^2$.

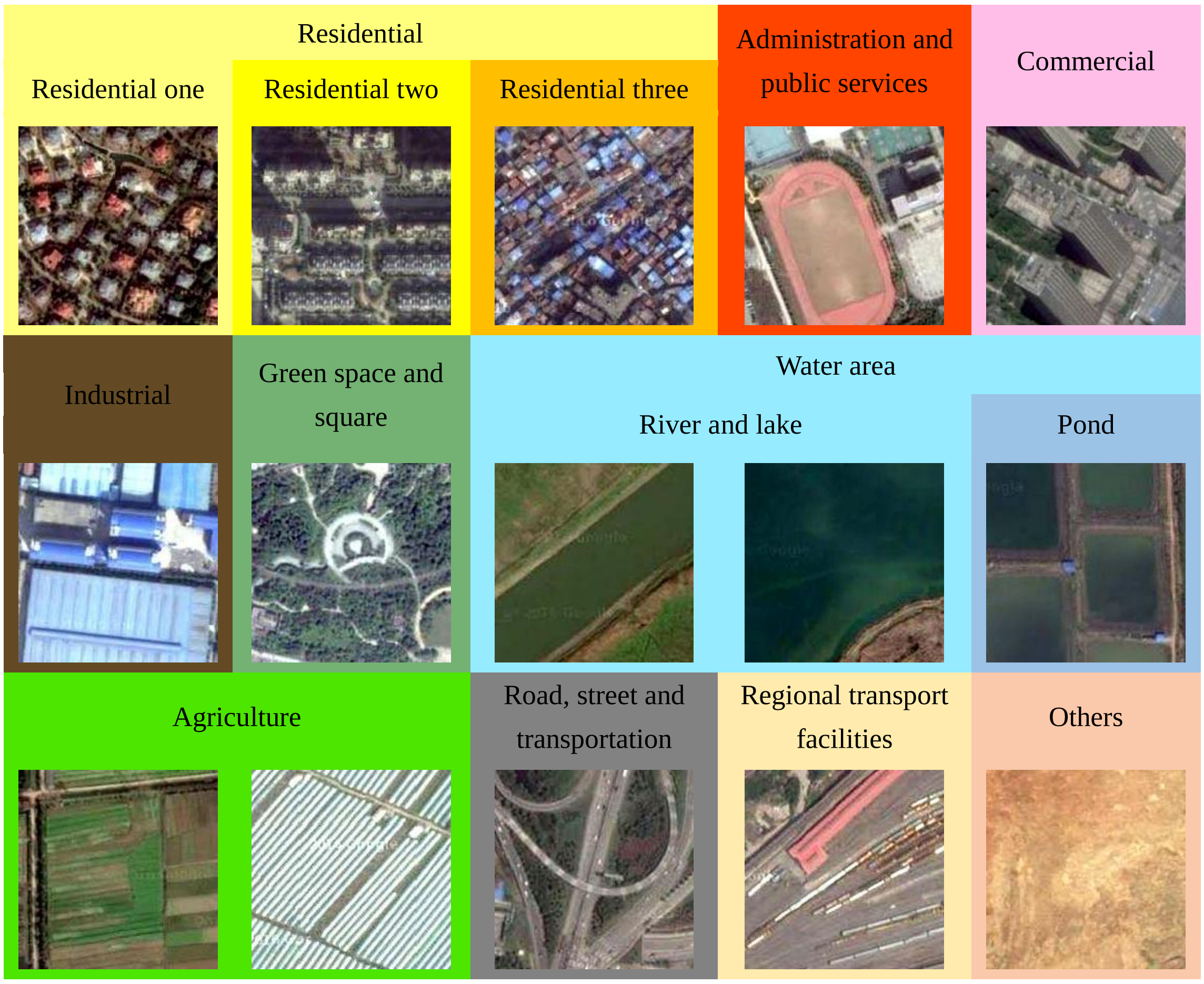

Fig. 2. The designed land-use type class hierarchy.

*C. Urban density data*

Accurate information on the urban density of Wuhan was obtained from building polygon data, which are building footprint maps of Wuhan from 2014. The original building footprint data were provided by the School of Urban Design, Wuhan University. Accurate urban density data were generated from the building polygon data. Two index of urban density were defined: building density and floor-area ratio. Firstly, the building polygon data were intersected by a grid. The grid size was tentatively set to 200 × 200 pixels (240 m × 240 m), which approximately corresponds to the same size as the land-use grid. The building density and floor-area ratio were calculated as follows:

$$BD(\mathrm{i},\mathrm{j}) = \frac{S_{building}(\mathrm{i},\mathrm{j})}{S_{Land}(\mathrm{i},\mathrm{j})}$$

$$FAR(\mathrm{i, j}) = \frac{S_{building}(\mathrm{i, j}) \times \mathrm{F(i, j)}}{S_{Land}(\mathrm{i, j})}$$

The pair (i, j) is the central location of the reference grid, $S_{building}$ indicates the building area in this grid, and $S_{Land}$ indicates the whole grid area. For the floor-area ratio, $\mathrm{F(i, j)}$ denotes the corresponding building floor in the reference grid.

*D. Population data*

The census data for 2014 were obtained from the School of Urban Design, Wuhan University. These data are available as an attribute of the polygons at the block level and cover the same area as the land-use data and urban density data. These data were divided into 1554 blocks, with the average area of a block being 1.6 $km^2$. The total population count was also available for each block. The blocks ranged from a maximum population of 66125 to a minimum of 0, and had a mean population of 5103 with a standard deviation of 5888. Accurate population density data were then generated from these population polygon data. As with the urban density data, the population polygon data were intersected by a grid. The grid size was set to 200 $\times$ 200 pixels (240 m $\times$ 240 m) and the corresponding area was the same as the urban density and land-use data. The population density was calculated from the population in each block and the corresponding area.

*E. Very high resolution (VHR) image data*

A very high spatial resolution remote sensing image with a 1.2 $m^2$ spatial resolution acquired by DigitalGlobe in January 2016 was used to evaluate the proposed method. This image covers an area of 47537 $\times$ 38100 pixels (2606 $km^2$), which consists of the central part of Wuhan. The image and GIS data were registered and matched so that they covered the same area. Because the images and the GIS data were acquired at different times, there were a few changes in land use and urban density. In Section 4, the potential changes of time differences are discussed in great detail.

## III. METHODS

To generate an accurate and informative mega-city analysis, three tasks have to be addressed: 1) land-use classification; 2) urban density estimation (building density estimation and floor-area ratio estimation); and 3) population estimation. As shown in Fig. 3, a CNN-based universal framework is proposed to solve these tasks. After the training of the CNN model, a detailed land-use classification map, building density estimation map, and floor-area ratio estimation map are generated, and the accuracy is evaluated for these three tasks. Subsequently, the land-use information, building density estimation, floor-area ratio estimation, and learned high-level features are combined to create a fine-scale population estimation within the study area. Finally, an accuracy evaluation of the produced population estimation map is undertaken with the ground truth dataset.

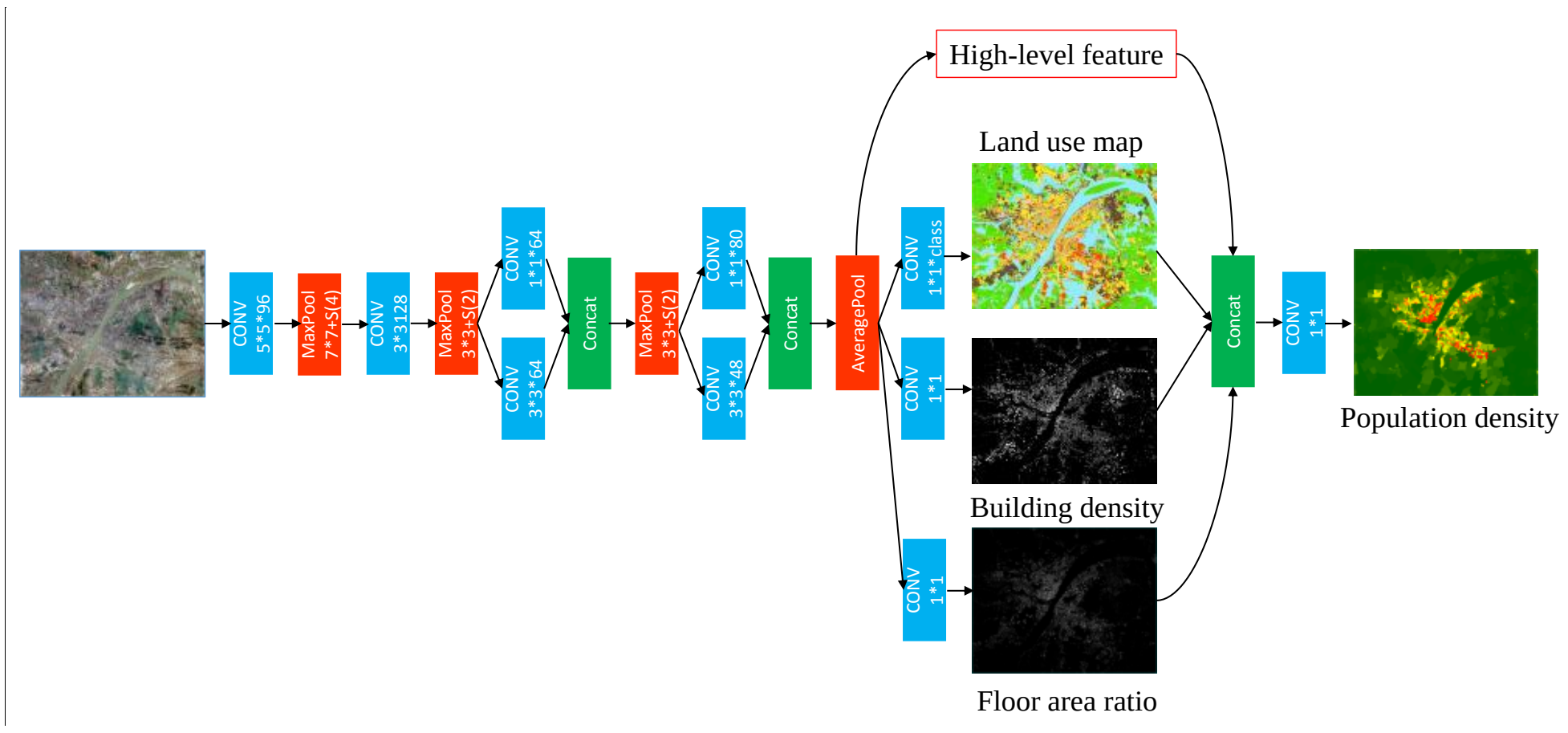

Fig. 3. The architecture of the proposed CNN framework.

### A. *Convolutional neural networks (CNNs)*

The CNN [30] [31] is a specialized kind of neural network for processing image data. They are made up of multiple feature extraction stages that have learnable weights and biases. We use three main types of layers to build a feature extraction stage: convolutional Layer, non-linear layer and pooling layer. In each stage, the original image or feature maps are convolved with learnable weights, and the extracted feature map are passed through some non-linear activation functions to form new feature maps, which are down-sampled via a pooling unit to generate output maps with a reduced spatial resolution.

**Convolutional layer:** The input to a convolutional layer is a set of feature map of size $m \times n \times r$ where $m$ and $n$ is the height and width of the feature map and $r$ is the number of channels (e.g., an image has $r = 3$). Each feature map is denoted as $x^i$. The convolutional layer include $k$ trainable filters of size $c \times c \times q$ and $k$ trainable bias parameters $b$. The trainable filters have a small receptive field than feature maps and also called the kernel $W$. The convolutional layer computes the output feature map $z^s = \sum_i^q W_i^s * x^i + b_s$, where $*$ is the 2D discrete convolution operator and $s$ indexes the filter number. The output is also a set of feature map of size $(m - c + 1) \times (n - c + 1) \times k$.

**Non-linear layer:** In a traditional CNN, this layer simply consists of a pointwise non-linear function applied to each component in the feature map. The non-linear layer computes the output feature map $a^s = f(\mathrm{z}^s)$, $f(\cdot)$ is commonly chosen to be a rectified linear unit (ReLU), and $f(\mathrm{x}) = \max(0, \mathrm{x})$.

**Pooling layer:** Pooling layer partitions the input feature maps into a set of non-overlapping or overlapping grid and executes a max operation over the grid of each feature map. The pooling layer is aim to reduce the spatial size of the feature map and decrease the number of parameters and computation in the network,

After the multiple feature extraction stages, the entire network is trained with back-propagation [25] of a supervised loss function such as the classic cross entropy of a softmax classifier output $\hat{\mathrm{y}}_i = soft\max(\mathrm{a}_i) = \frac{e^{a_i}}{\sum_j^{N_c} e^{a_j}}$, where $\hat{\mathrm{y}}_i$ is the activations of the previous layer node $i$ pushed through a softmax function. The target output $y$ is represented as a $1 - of - K$ vector, where $K$ is the number of outputs and $L$ is the number of layers:

$$J(\theta) = -\sum_i^{N_c} y_i \log(\hat{\mathrm{y}}_i) + \lambda \sum_l^L (\| \mathrm{W}^{(l)} \|^2)$$

where $\hat{y}_i$ is the activations of the previous layer, $\lambda$ is a regularization term (also called a weight decay term), $\{W, b\} \in \theta$ and $l$ indexes the layer number. Our goal is to minimize $J(\theta)$ as a function of $W$ and $b$. To train the CNN, we apply stochastic gradient descent with back-propagation to optimize the function.

Typical CNNs, including LeNet [30], AlexNet [25], and its deeper successors [23] [32] [33], ostensibly need fixed-sized images and generate non-spatial outputs. The input of fully-connected layers need fixed dimensions and discard the spatial information. However, for the urban land-use classification, urban density, or population density estimation tasks, the size of the input image is not fixed. For various sizes of image, we can view these fully-connected layers as convolutional layers with the size of each kernel have covered the entire input feature map. By changing the fully-connected layers into convolutional layers, we can get a fully convolutional neural network which only contain the convolutional layers and handle the various sizes of image.

*B. Overall architecture*

Taking advantage of hierarchical feature learning and end-to-end learning processing, training a deep CNN model is simple and efficient. Compared to most of the traditional methods [6]-[9], which need several elaborate manually designed parts, our CNN model can be automatically trained from the massive data and generate different-level products for our tasks.

In this study, we propose a novel CNN-based universal framework to learn a hierarchical feature representation from the massive data and simultaneously generate a land-use map, an urban density estimation, and a population density estimation. The detailed network architecture is outlined in Table III, and we also visualize the network architecture in Fig. 3. The convolutional layer parameters are denoted as "Conv(receptive field size)−(number of features)". The pooling layer parameters are denoted as “Maxpool(pooling region size)−(pooling stride)”.

As Table III shows, the CNN model contains four feature extraction stages and two task-dependent layers. In order to consider the specific attributes of our task and the remote sensing data, we designed these two task-dependent layers. For task-dependent layer I, we designed three different softmax functions for land-use classification, building density estimation, and floor-area ratio estimation. We can also consider the three different tasks as three subnets which share the same network architecture. The population estimation task is more complex and is highly correlated with socioeconomic factors, such as land use, urban density, and built-up area fraction. We therefore designed task-dependent layer II to combine the information learned by the previous layer and assemble our network architecture into a universal framework for the population estimation.

TABLE III

THE ARCHITECTURE OF THE PROPOSED CNN MODEL

| CNN configuration | | | |
|---|---|---|---|
| Input image | | | |
| Feature extraction I | Conv5*5-96 | | |
| | Maxpool 7*7-4 | | |
| Feature extraction II | Conv3*3-128 | | |
| | Maxpool 3*3-2 | | |
| Feature extraction III | Conv1*1-64 | Conv3*3-64 | |
| | Feature concatenation | | |
| | Maxpool 3*3-2 | | |
| Feature extraction IV | Conv1*1-80 | Conv3*3-48 | |
| | Feature concatenation | | |
| | Avepool 10*10-1 | | |
| Task-dependent layer I | Subnet-A: Land use classification<br>Conv1*1-13<br>Softmax classifier | Subnet-B: BD estimation<br>Conv1*1-25<br>Softmax classifier | Subnet-C: FAR estimation<br>Conv1*1-32<br>Softmax classifier |
| | Output concatenation with Avepool feature in Feature extraction Stage IV | | |
| Task-dependent layer II | Population estimation<br>Conv1*1-40<br>Softmax classifier | | |

### C. Details of the parameter settings

Based on the designed mega-city analysis tasks, the most critical task in the deep learning method is the collection of a large volume of labeled data for training the deep model. By taking advantage of the remote sensing image and the corresponding GIS data, it is both easy and efficient to collect the large volume of training data.

For the land-use task, we manually selected 26363 training images with a size of 200 × 200 pixels (240 m × 240 m) and their corresponding labels.

For the urban density task, we manually selected 75733 training images with a size of 200 × 200 pixels (240 m × 240 m) and their corresponding building density values and floor-area ratio values. Based on the urban density data in Wuhan, the building density and floor-area ratio values are continuous data between 0–1 and 0–10. For simplicity, we equally discretized the building density values to 25 levels between 0–1 and equally discretized the floor-area ratio values to 32 levels between 0–10.

For the population density task, we manually selected 98219 training images with a size of 200 × 200 pixels (240 m × 240 m) and their corresponding population number. Based on the census data from Wuhan, the range of the population is between 0–7500. As in the urban density task, we also equally discretized the population number to 40 levels between 0–7500.

We trained the models using stochastic gradient descent with a batch size of 64 examples, momentum of 0.9, and weight decay of 0.0005. The learning rate was initialized as 0.01 and reduced every epoch by 0.95. The experiments were run on a PC with an NVIDIA Titan GPU with 6 GB of memory and a single Intel core i7 CPU. The operating system was Ubuntu 14.04, and the implementation environment was under a modified version of the MXNet toolbox [34].

*D. End-to-end training*

*1) Loss function*

We combine the four tasks to train our CNN model: Land use classification, BD estimation, FAR estimation and Population estimation. The loss function of the entire CNN model was defined as below:

$$J(\theta) = J_{Land}(\theta_1) + J_{BD}(\theta_2) + J_{FAR}(\theta_3) + J_{Pop}(\theta_4 \mid \theta_1, \theta_2, \theta_3)$$

$J_{Land}(\theta_1)$ is the learning objective for land use classification task, we formulated as a multi-class classification problem. We use the cross-entropy loss $J_{Land}(\theta_1) = -\sum_{i}^{N_c} y_i \log(\hat{y}_i) + \lambda \sum_{l}^{L} (\| W^{(l)} \|^2)$ to define the objective function. For the land use classification task, the class number $N_c$ is 13.

Similar to land use classification task, BD estimation $J_{BD}(\theta_2)$ and FAR estimation $J_{FAR}(\theta_3)$ are also formulated as a multi-class classification problem. For the BD estimation, we equally discretized the building density values to 25 levels between 0–1, so the class number $N_c$ is 25. For the FAR estimation, we equally discretized the building density values to 32 levels between 0–10, so the class number $N_c$ is 32.

For the population estimation task, the loss function is unlike traditional CNN model, because the loss term of the population estimation depends on the output of the Land use classification, BD estimation and FAR estimation. So, based on the chain rule of back-propagation, the gradient of $J_{Pop}(\theta_4 \mid \theta_1, \theta_2, \theta_3)$ involves the gradients of $J_{Land}(\theta_1)$, $J_{BD}(\theta_2)$ and $J_{FAR}(\theta_3)$. Because we also equally discretized the population number to 40 levels between 0–7500, the population estimation task are also formulated as a multi-class classification problem like before. The class number $N_c$ is 40 for population estimation task.

*2) 2-stage training*

Because all the tasks shared the same feature extraction stage, we therefore need to develop a technique that allows for sharing feature extraction stages between the different networks, rather than learning several separate networks. In this paper, we adopt a pragmatic 2-step training algorithm to train the multi-task CNN model via alternating optimization.

**First-step:** Since we employ different task in different networks, there are different types of training images in the learning process, such as land use, building density, and floor-area-ratio. In this case, some of the loss functions are not used. For example, for the sample of land use, we only compute $J_{Land}(\theta_1)$, and the others losses are set as 0. This can be implemented directly with a sample type indicator. Then the overall learning target can be formulated as:

$$J_{First}(\theta) = \beta_1 J_{Land}(\theta_1) + \beta_2 J_{BD}(\theta_2) + \beta_3 J_{FAR}(\theta_3)$$

where $\beta \in \{0,1\}$ is the sample type indicator. Since the loss term of the population estimation depends on the output of the Land use classification, BD estimation and FAR estimation, it is difficult to directly training the whole network from random initialization, we only optimized the Land use classification, BD estimation and FAR estimation in first step.

**Second-step:** After the first step training, we added the $J_{Pop}(\theta_4 \mid \theta_1, \theta_2, \theta_3)$ to the overall learning target:

$$J_{Second}(\theta) = \beta_1 J_{Land}(\theta_1) + \beta_2 J_{BD}(\theta_2) + \beta_3 J_{FAR}(\theta_3) + \beta_4 J_{Pop}(\theta_4 \mid \theta_1, \theta_2, \theta_3)$$

Similar to first step, $\beta \in \{0,1\}$ is the sample type indicator. In second step, we decrease the learning rate to 0.001 for the previous layers. For the Task-dependent layer II, the learning rate was set as 0.01.

According to the 2-stage training, both networks share the same feature extraction stage and form a unified network.

## IV. RESULTS

### *A. Land-use mapping*

#### *3) Evaluation of the land-use accuracy*

The study area of 2606 $km^2$ covers the center of Wuhan and the surrounding regions. The generated map of the land use for Wuhan is shown in Fig. 4. Within the large city area, farmland accounts for the largest amount, covering 670.8 $km^2$ (25.74%) of the image. From Fig. 4, we can see that agriculture is the dominant land use in the surrounding regions. As a lake city, the water area covers 611.7 $km^2$ (23.48%) and is the second largest land-use type in Wuhan. As Fig. 4 shows, the proposed CNN model can generate an accurate result, even for the branches of the Yangtze River. Among the residential areas, most of the regions are classified as “Residential two” (10.24%). Consider the “villages in the city” problem, large areas belong to “Residential three”, with relatively poor living conditions, even in the central area of Wuhan. As the most important economic center in central China, 384.8 $km^2$ (14.77%) is classified as industrial area, which surrounds the central area of Wuhan. In addition to the obvious patterns, such as water bodies and the different levels of residential type, the CNN model also extracts the major roads and regional transport facilities. The resulting land-use map discovers obvious structures composed of clusters of the same land-use type, which suggests that the proposed model is suitable for the large-scale monitoring of urban areas. Moreover, the land-use ratio (Fig. 6) also reflects the global patterns by integrating the qualitative and quantitative information.

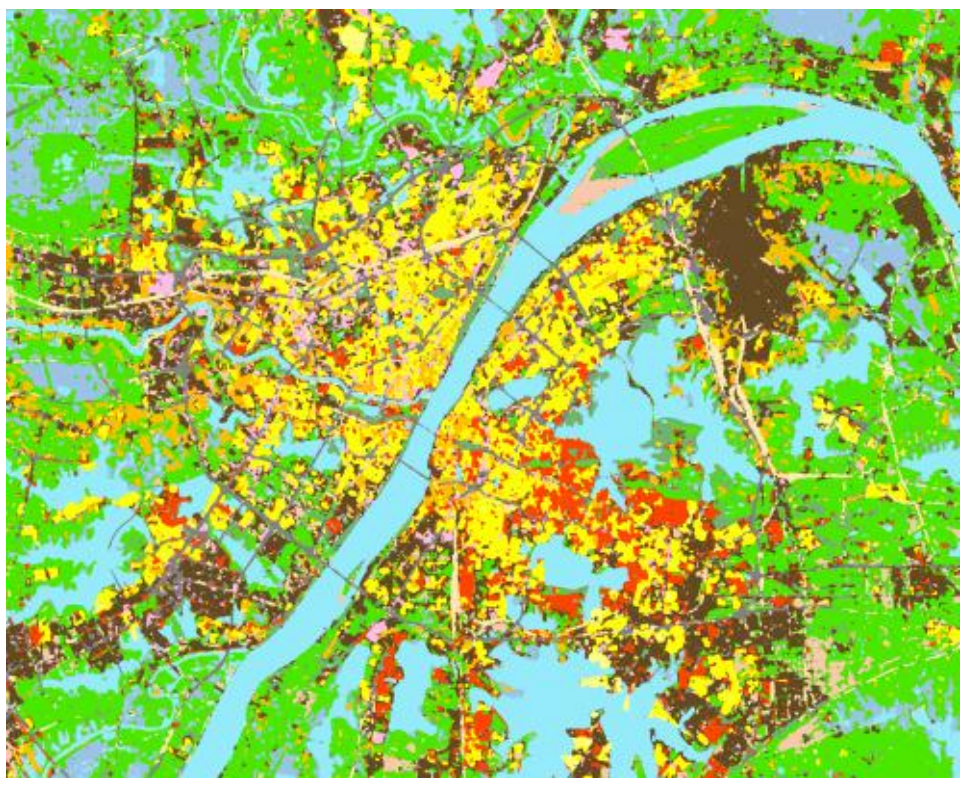

(a)

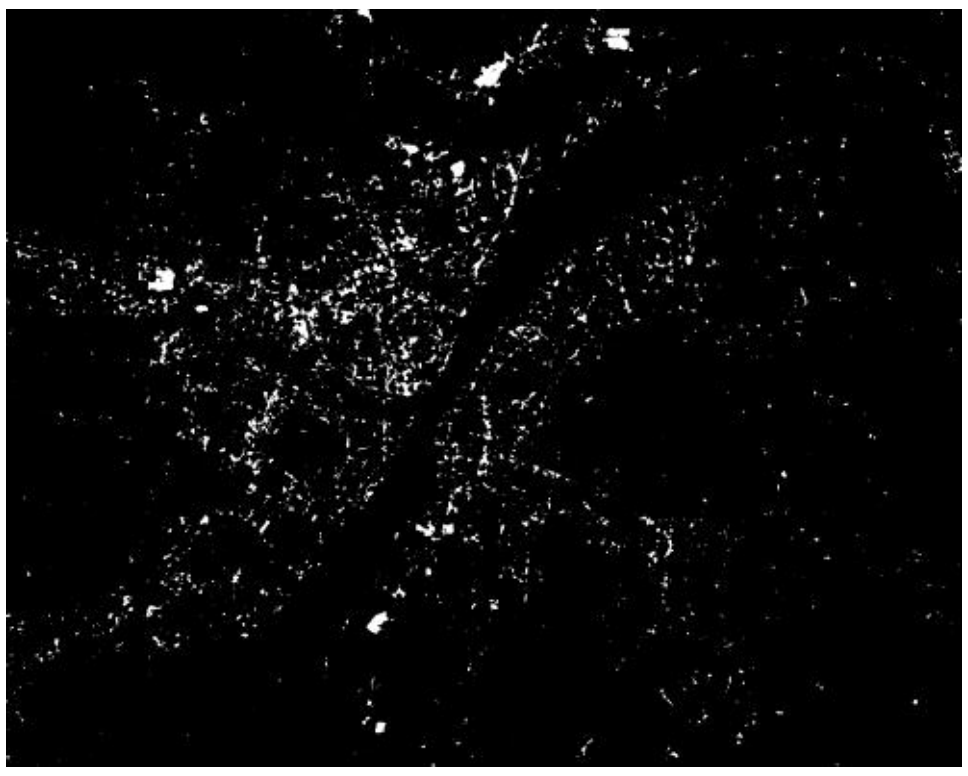

(b)

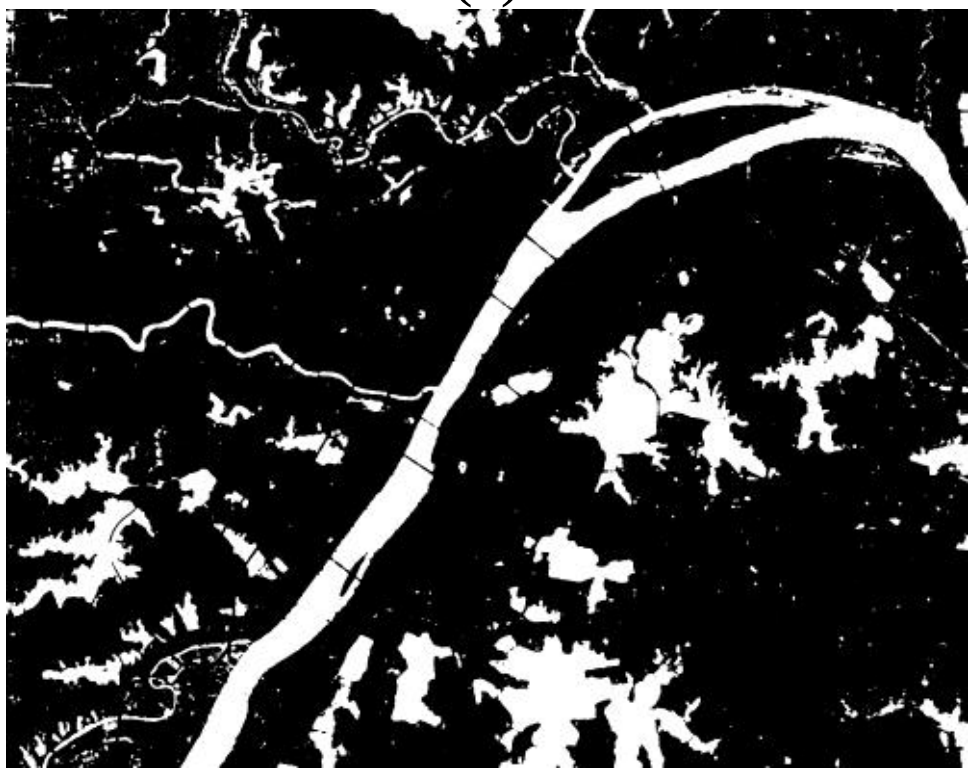

(c)

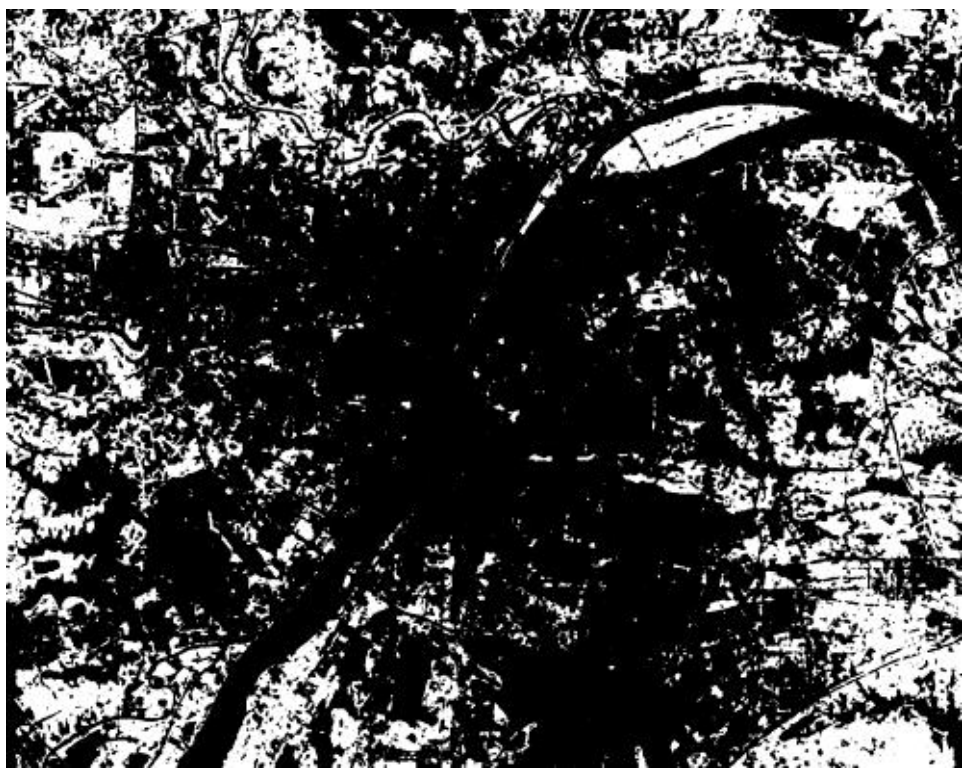

(d)

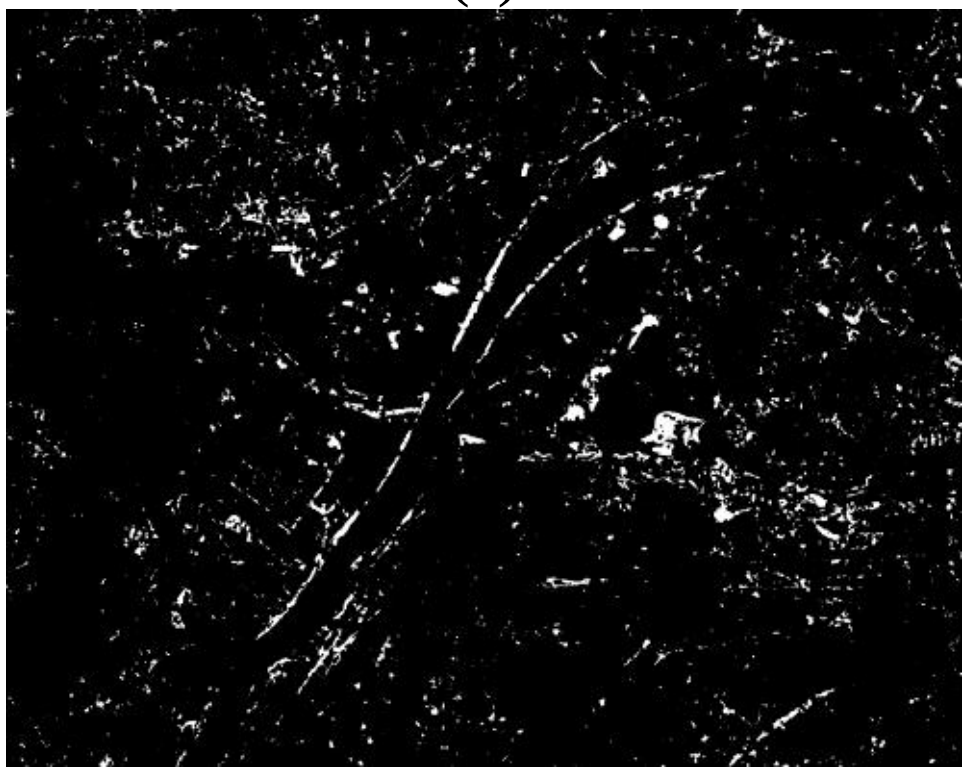

(e)

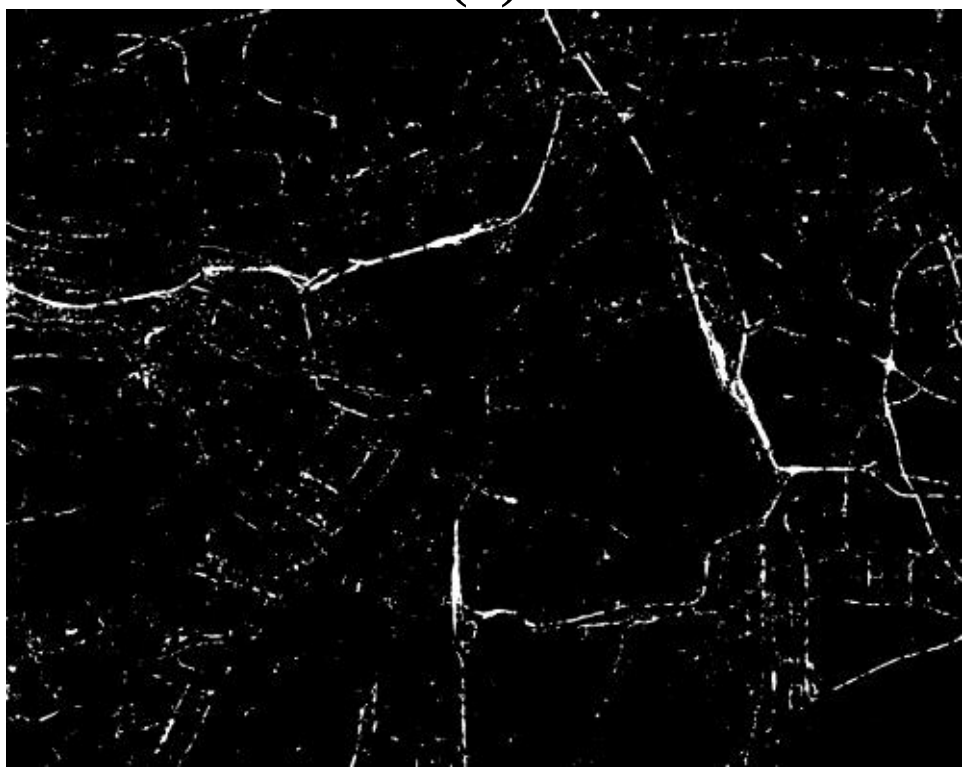

(f)

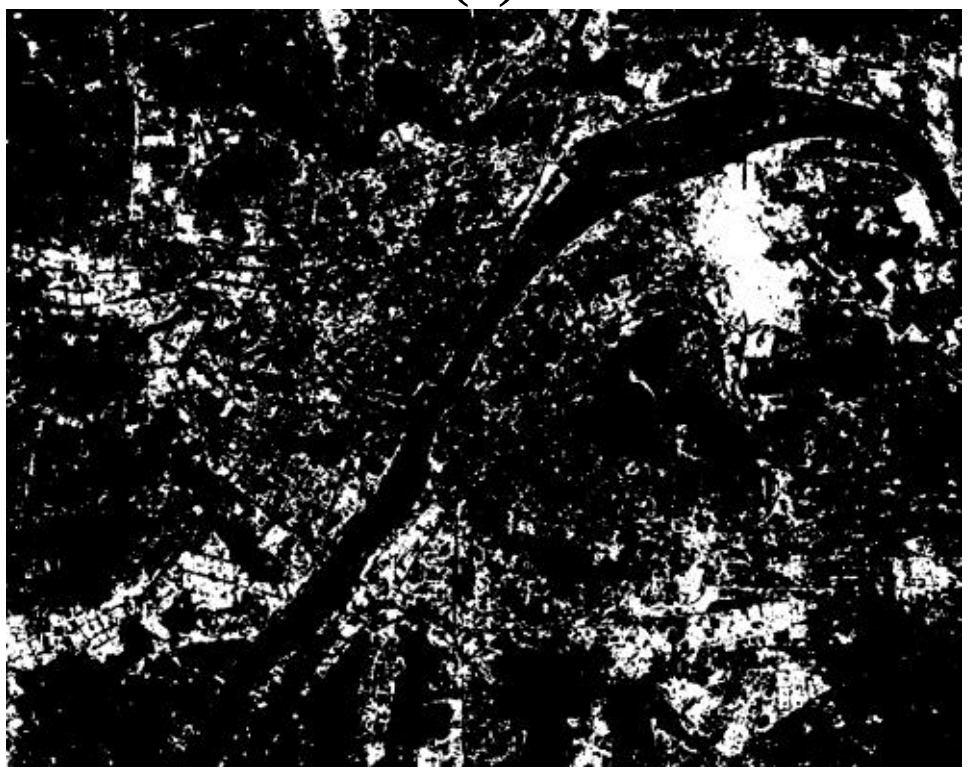

(g)

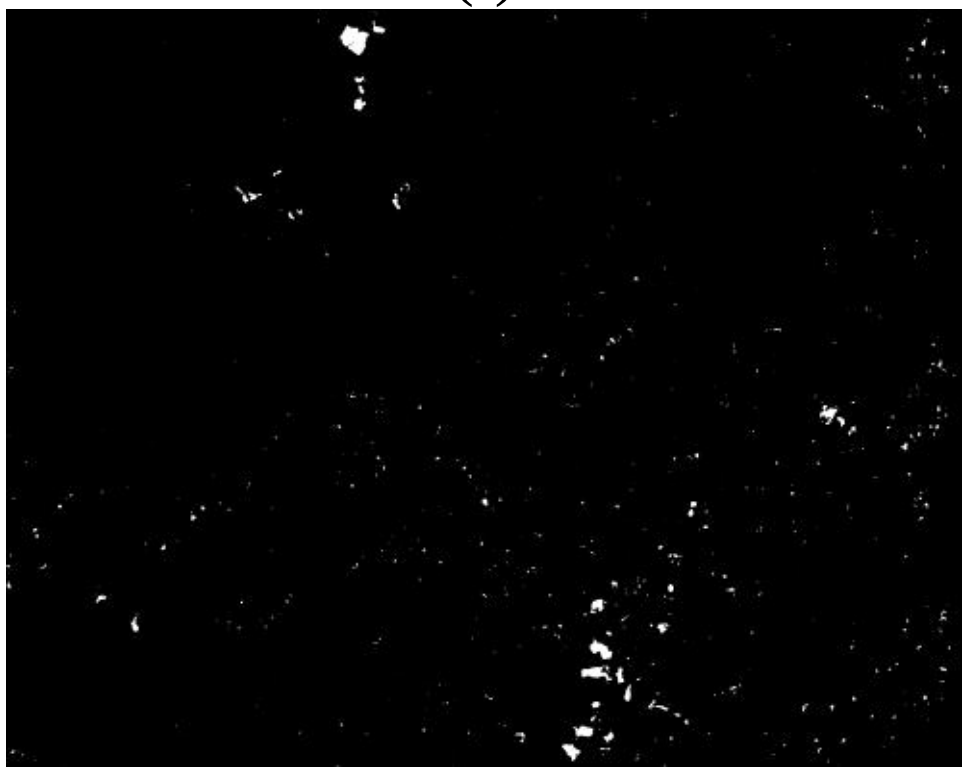

(h)

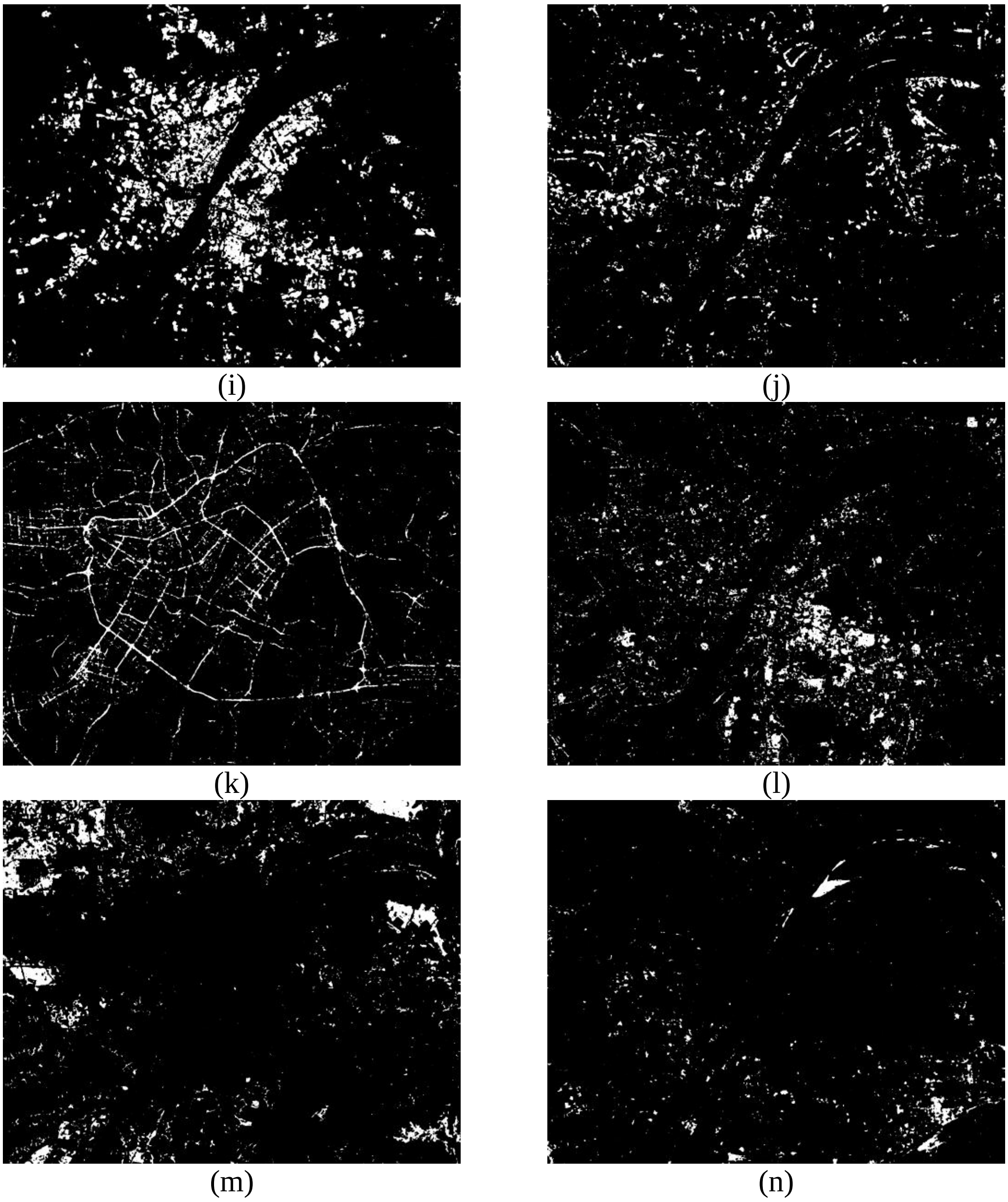

Fig. 4. (a) The derived land-use type classification for the city of Wuhan. (b) Commercial, (c) Water area – river and lake, (d) Agriculture, (e) Green space and square, (f) Regional transport facilities, (g) Industrial, (h) Residential one, (i) Residential two, (j) Residential three, (k) Road, street and transportation, (l) Administration and public services, (m) Water area – pond, (n) Others.

We evaluated the classified land use via an error matrix based on 5271 randomly sampled images, with at least 50 images for each category. As shown in Table IV, error matrices were generated to calculate the overall accuracy (OA), the user’s and producer’s accuracies, and the Kappa statistics. The land-use classification resulted in an OA of 92.35% and a Kappa index of 0.9143. It can be observed from Table IV that, in most land-use types, with the exception of the “Others” land-use type, the accuracies reach 80% or higher. The classification of “Water area – pond” obtains the lowest user’s accuracy due to the high similarity to “Water area – river and lake” and the misclassifications regarding farmland, which is the dominant background land-use type in the rural areas. The good classification results and the reasonable OA indicate the benefit of using a deep learning method to learn from massive data and automatically find the relationship between the data and the task.

TABLE IV

ACCURACY ASSESSMENT OF THE GENERATED LAND-USE MAP

| | | Reference | | | | | | | | | | | | | | |
|---|---|---|---|---|---|---|---|---|---|---|---|---|---|---|---|---|
| | | 1 | 2 | 3 | 4 | 5 | 6 | 7 | 8 | 9 | 10 | 11 | 12 | 13 | Total | UA [%] |
| Classification | 1 | **198** | 1 | 1 | 2 | 0 | 3 | 1 | 1 | 0 | 1 | 1 | 0 | 0 | 209 | **94.74** |
| | 2 | 0 | **846** | 2 | 7 | 0 | 0 | 0 | 0 | 0 | 0 | 0 | 4 | 0 | 859 | **98.49** |
| | 3 | 1 | 4 | **621** | 3 | 5 | 3 | 0 | 0 | 0 | 0 | 0 | 11 | 0 | 648 | **95.83** |
| | 4 | 0 | 7 | 12 | **363** | 0 | 1 | 0 | 0 | 0 | 0 | 3 | 0 | 0 | 386 | **94.04** |
| | 5 | 0 | 4 | 9 | 5 | **342** | 6 | 0 | 0 | 0 | 18 | 2 | 0 | 0 | 386 | **88.60** |
| | 6 | 15 | 0 | 13 | 3 | 7 | **839** | 2 | 3 | 0 | 1 | 33 | 0 | 1 | 917 | **91.49** |
| | 7 | 0 | 0 | 0 | 1 | 1 | 0 | **139** | 4 | 1 | 0 | 0 | 0 | 0 | 146 | **95.21** |
| | 8 | 0 | 0 | 0 | 0 | 0 | 2 | 16 | **463** | 2 | 0 | 11 | 0 | 0 | 494 | **93.72** |
| | 9 | 1 | 0 | 4 | 5 | 0 | 3 | 2 | 2 | **247** | 0 | 1 | 0 | 0 | 265 | **93.21** |
| | 10 | 2 | 8 | 8 | 3 | 16 | 0 | 0 | 0 | 0 | **197** | 0 | 1 | 0 | 235 | **83.83** |
| | 11 | 1 | 0 | 10 | 13 | 0 | 23 | 3 | 0 | 0 | 2 | **414** | 2 | 0 | 468 | **88.46** |
| | 12 | 0 | 17 | 12 | 0 | 0 | 0 | 0 | 0 | 0 | 0 | 0 | **136** | 0 | 165 | **82.42** |
| | 13 | 0 | 2 | 25 | 0 | 1 | 1 | 0 | 1 | 0 | 0 | 0 | 0 | **63** | 93 | **67.74** |
| Total | | 218 | 889 | 717 | 405 | 372 | 881 | 163 | 474 | 250 | 219 | 465 | 154 | 64 | **5271** | |
| PA [%] | | **90.82** | **95.16** | **86.61** | **89.63** | **91.93** | **95.23** | **85.28** | **97.68** | **98.80** | **89.95** | **89.03** | **88.31** | **98.44** | **OA [%]** | **92.35** |

1) Commercial, 2) Water area – river and lake, 3) Agriculture, 4) Green space and square, 5) Regional transport facilities, 6) Industrial, 7) Residential one, 8) Residential two, 9) Residential three, 10) Road, street and transportation, 11) Administration and public services, 12) Water area – pond, 13) Others.

*4) Change analysis*

Because the very high spatial resolution remote sensing data (acquired in January 2016) and the original land-use data (acquired in 2014) were obtained at different times, there were a few changes between the land-use map derived by the proposed CNN model and the original land-use data provided by the Wuhan Land Resources and Planning Bureau for 2014 (Fig. 5).

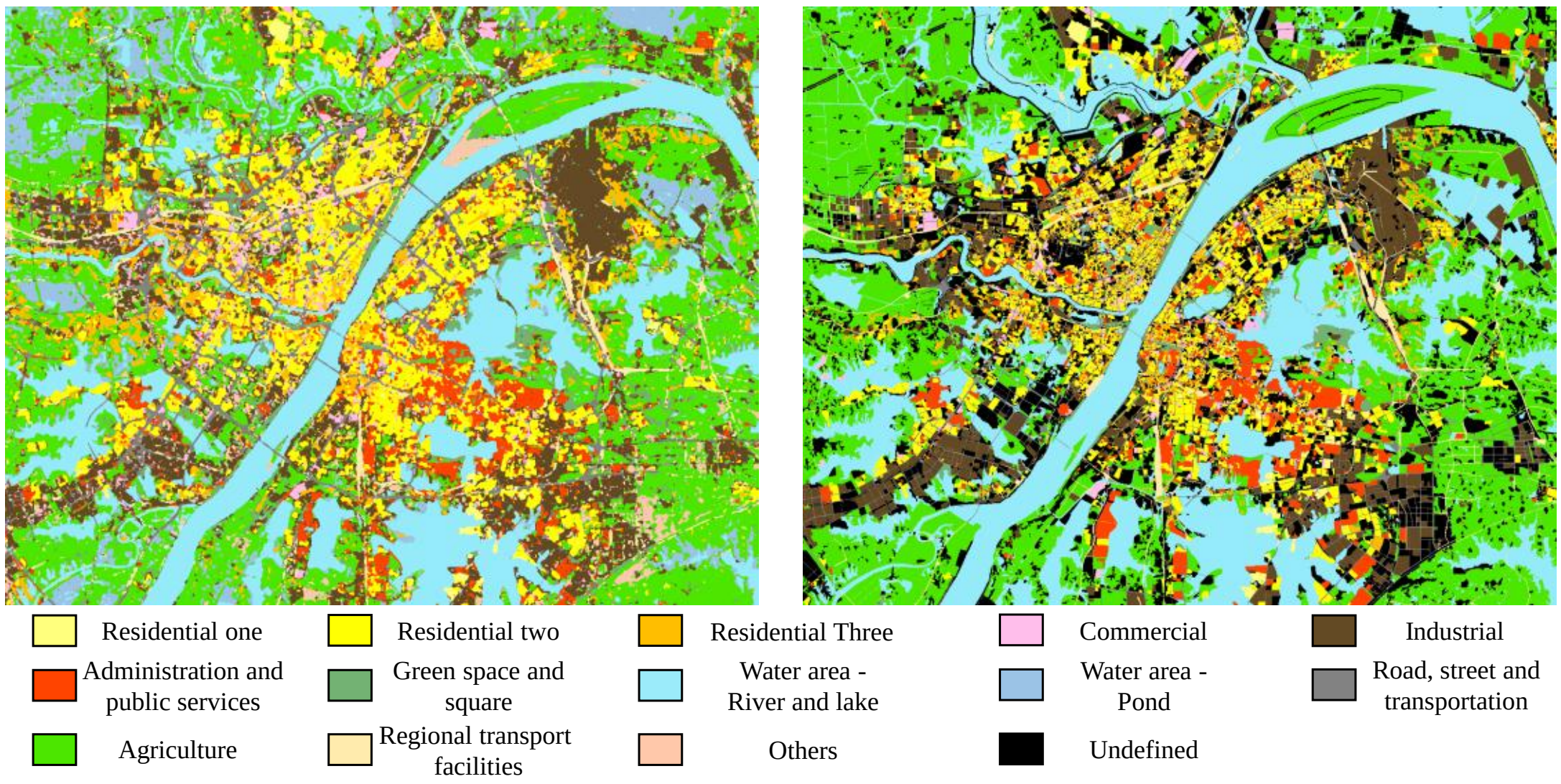

Fig. 5. Comparison between the land-use types. Left: the derived land-use type classification for the city of Wuhan in 2016. Right: the original land-use data for the city of Wuhan from 2014.

The land-use classification results for 2016 and the original land-use data from 2014 were used to compare the land-use types of Wuhan in terms of ratios and spatial distribution. As shown in Fig. 5, compared to the original land-use data from 2014, the derived land-use map for 2016 reveals basically the same patterns, and the land-use spatial distribution remains the same. For the land-use ratio of the different categories (Fig. 6), we compared the nine main classes and we did not consider the "Others" class. For residential, we separated it into the three subclasses, allowing a better analysis of the living conditions of Wuhan.

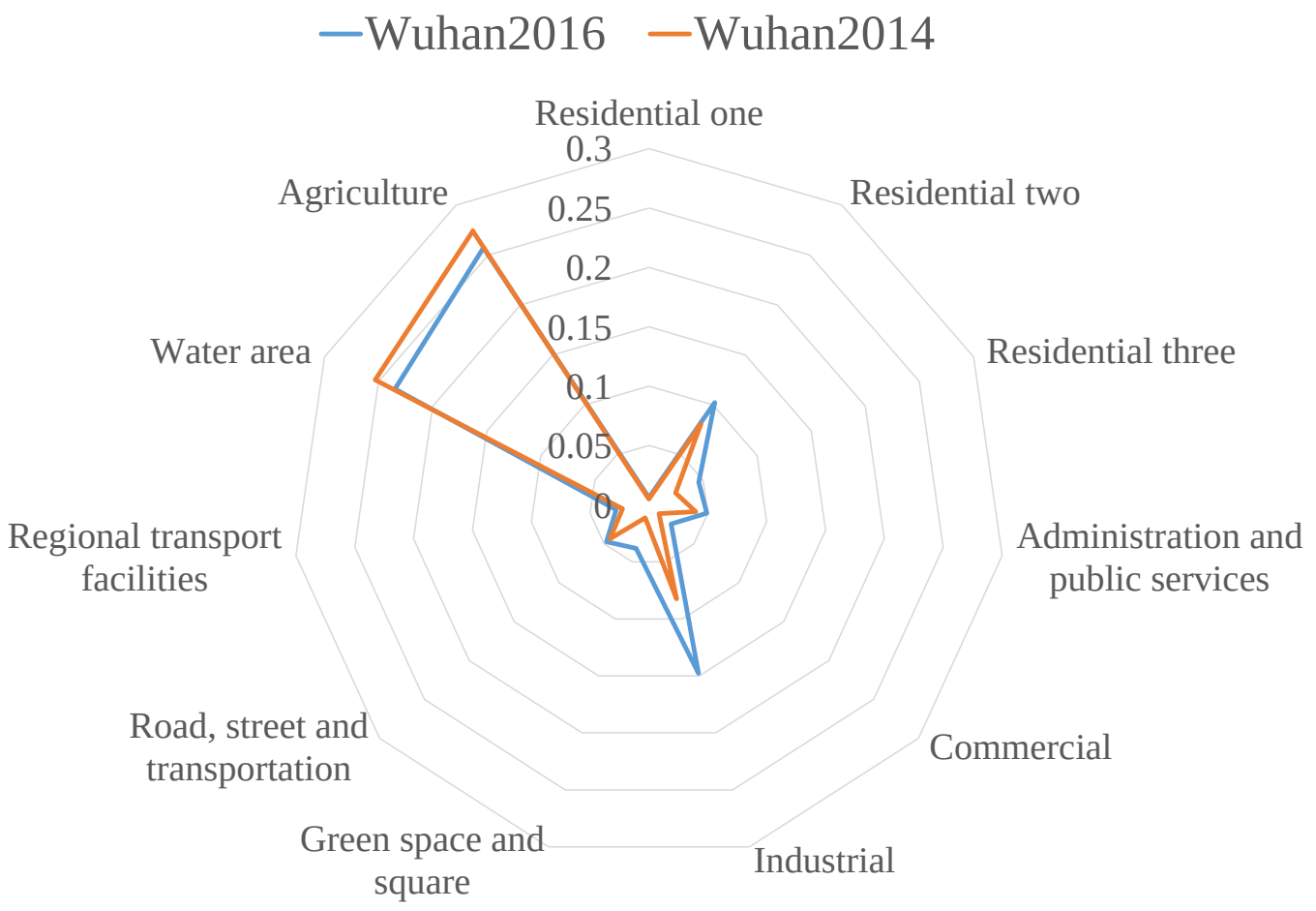

Fig. 6. The land-use ratios of the different categories in Wuhan 2016 and Wuhan 2014.

As shown in Table V and Fig. 6, according to the rapid urbanization, we can observe that there has been a significant decrease in the water areas and agriculture and a rapid increase in the industrial and commercial areas compared to 2014. As the educational and transportation center of central China, Wuhan has also witnessed a dramatic development in administration and public services, such as the university campus and railway network. However, the residential condition, or the so-called "villages in the city", is still a severe problem. In the central area of Wuhan, there still remains a large proportion of the low-level residential type (Fig. 5).

TABLE V

RATIO COMPARISON OF THE LAND-USE TYPES BETWEEN WUHAN 2016 AND WUHAN 2014

| Land-use type | 1 | 2 | 3 | 4 | 5 | 6 | 7 | 8 | 9 | 10 | 11 |
|---|---|---|---|---|---|---|---|---|---|---|---|
| Wuhan 2016(%) | 2.44 | 23.48 | 25.74 | 3.81 | 2.78 | 14.76 | 0.60 | 10.24 | 4.59 | 4.72 | 4.92 |
| Wuhan 2014(%) | 1.11 | 25.30 | 27.41 | 1.12 | 2.24 | 8.24 | 0.48 | 8.08 | 2.44 | 4.34 | 3.97 |

1) Commercial, 2) Water area, 3) Agriculture, 4) Green space and square, 5) Regional transport facilities, 6) Industrial, 7) Residential one, 8) Residential two, 9) Residential three, 10) Road, street and transportation, 11) Administration and public services.

### *B. Urban density derivation*

Two index of urban density were used: building density and floor-area ratio. The accuracy of the derived building density and floor-area ratio maps was assessed via the urban density data of Wuhan from 2014 provided by the School of Urban Design, Wuhan University. Because the very high spatial resolution remote sensing data (acquired in January 2016) and the urban density data (acquired in 2014) were obtained at different times, for a better analysis, we also used Google Earth to visually interpret the difference between the urban density map derived by the CNN model and the original urban density data of Wuhan from 2014.

The resulting map of building density for the city of Wuhan is shown in Fig. 7. The major difference between Wuhan in 2016 and Wuhan in 2014 is mainly found in the surrounding regions. Due to the rapid development and construction, a lot of farmland and other land-use types have change into industrial, residential, or commercial, which is consistent with the land-use ratio change. In the central area of Wuhan, the change of building density is marginal due to the lack of large free areas for construction.

Additionally, the per pixel correlation between Wuhan in 2016 and Wuhan in 2014 was assessed, showing reasonable results. As is shown in Fig. 8, the mean absolute error of the derived and actual building density is 0.023 and the correlation is 0.8404. Due to the building density change between 2014 and 2016, we can also observe the difference, which mainly indicates the building structure or land-use change.

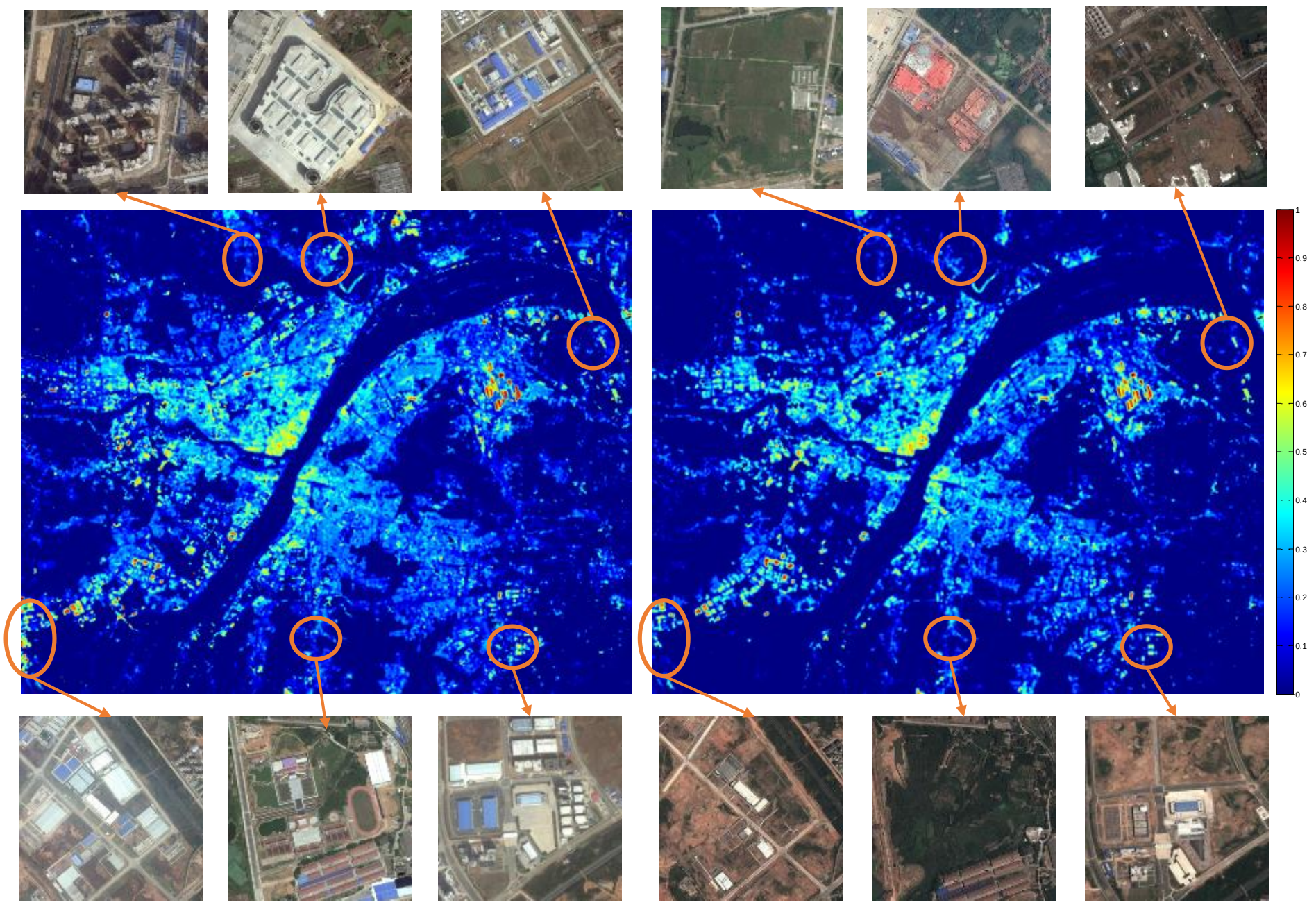

Fig. 7. Comparison between the building density. Left: the derived building density map for the city of Wuhan in 2016. Right: the original building density data for the city of Wuhan from 2014.

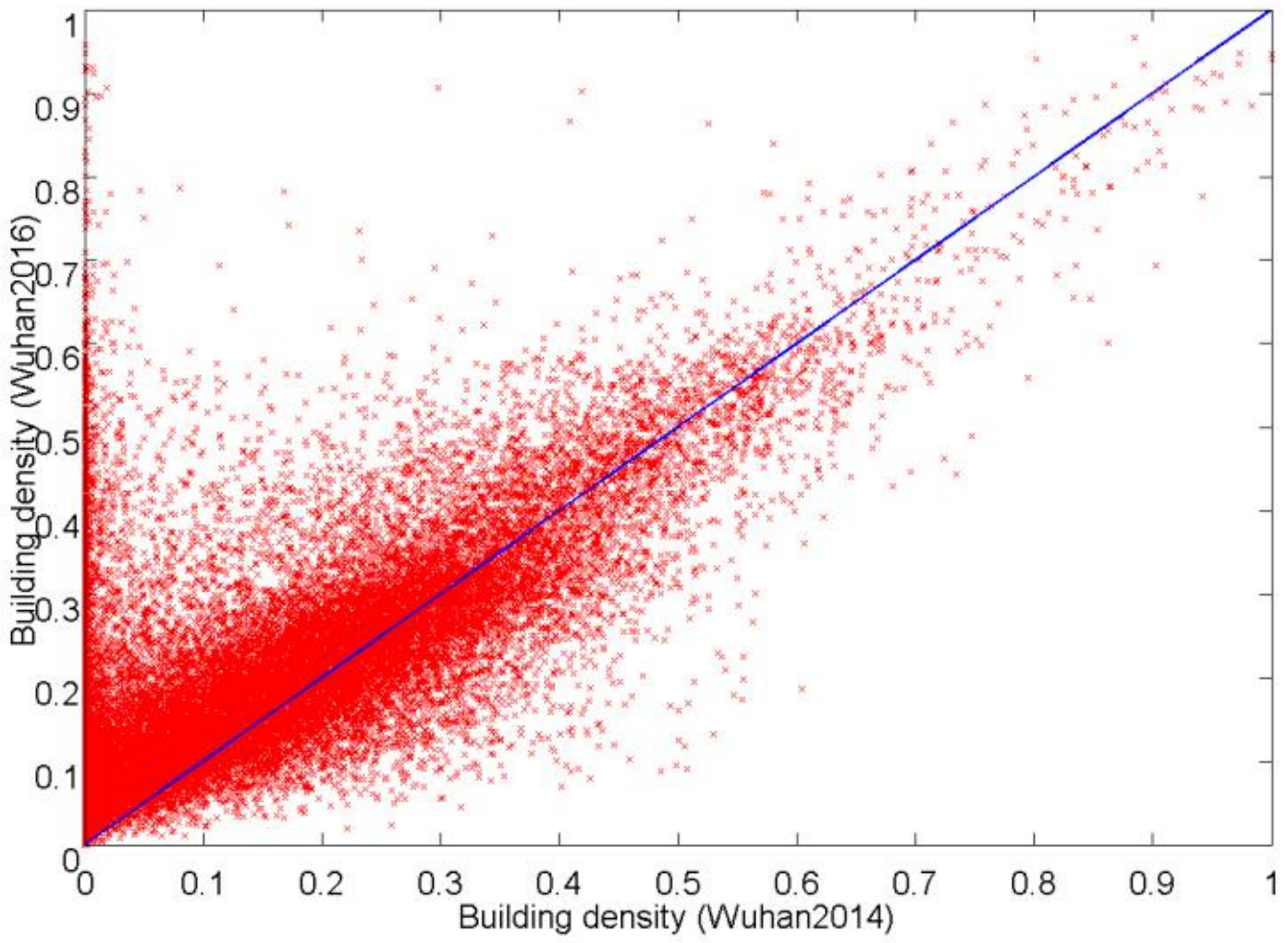

Fig. 8. Building density correlation scatterplot between Wuhan 2016 and Wuhan 2014.

The resulting map of the floor-area ratio for the city of Wuhan is shown in Fig. 9. Compared to the building density, the floor-area ratio mainly reflects the tall buildings or high-rise apartments, which gives us another factor to analyze the city density. For the different levels of residential type, the mean building density for “Residential” three” is higher than for “Residential two”, but with a lower mean floor-area ratio (Table VI). The commercial areas have the highest mean building density and second largest mean floor-area ratio, which reveals that the commercial areas have more dense building structures and taller buildings than the other land-use types.

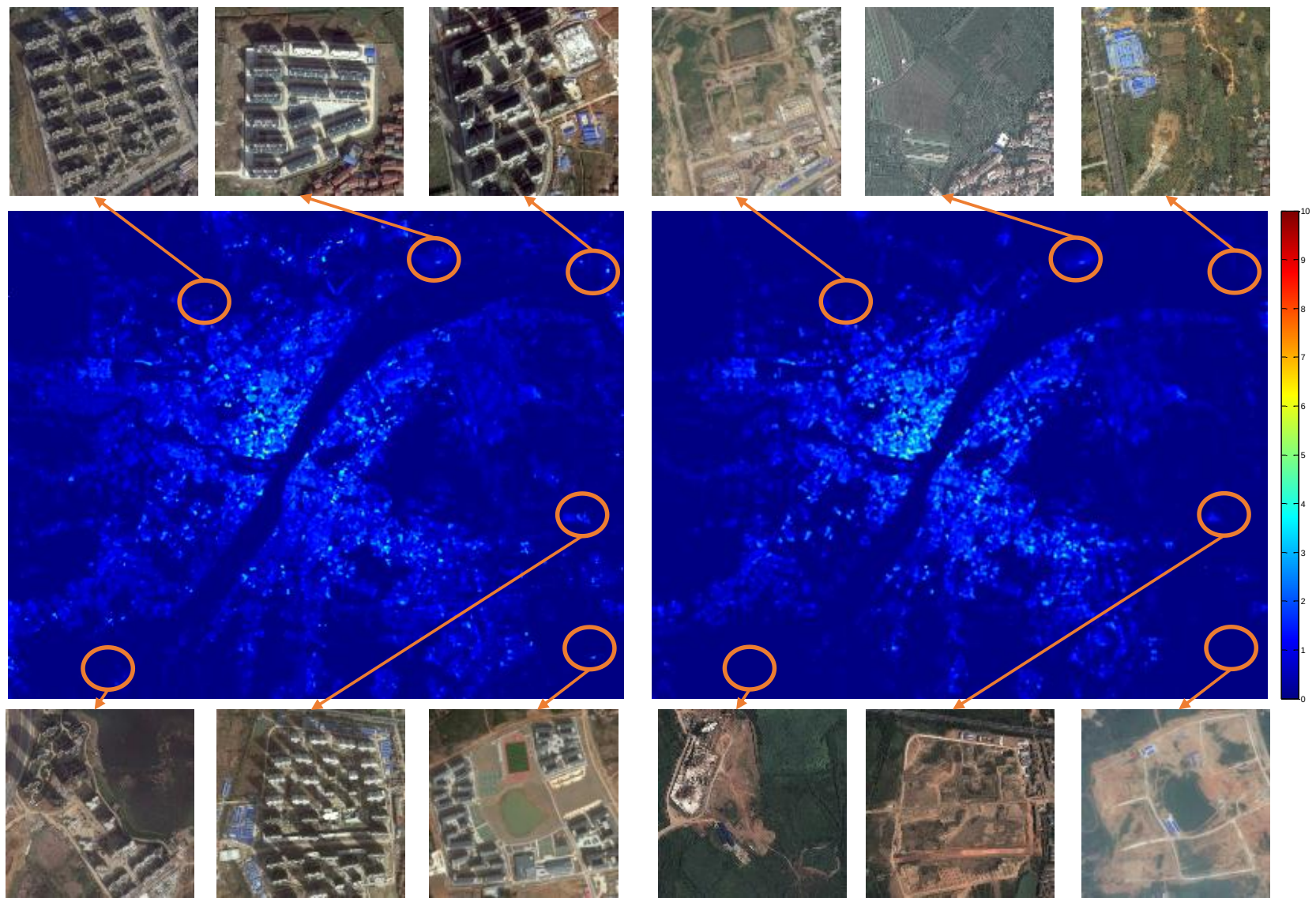

Fig. 9. Comparison between the floor-area ratio. Left: the derived floor-area ratio map for the city of Wuhan in 2016. Right: the original floor-area ratio data for the city of Wuhan from 2014.

TABLE VI

COMPARISON BETWEEN THE MEAN BUILDING DENSITY AND MEAN FLOOR-AREA RATIO

| Land use type | | 1 | 2 | 3 | 4 | 5 | 6 | 7 | 8 | 9 | 10 | 11 |
|---|---|---|---|---|---|---|---|---|---|---|---|---|
| Wuhan 2016 | BD | 0.29 | 0.03 | 0.06 | 0.08 | 0.12 | 0.21 | 0.17 | 0.24 | 0.26 | 0.13 | 0.17 |
| | FAR | 0.88 | 0.09 | 0.16 | 0.24 | 0.33 | 0.50 | 0.54 | 1.29 | 0.66 | 0.43 | 0.62 |
| Wuhan 2014 | BD | 0.25 | 0.008 | 0.02 | 0.05 | 0.07 | 0.21 | 0.16 | 0.21 | 0.28 | 0.14 | 0.17 |
| | FAR | 0.93 | 0.01 | 0.03 | 0.20 | 0.15 | 0.38 | 0.69 | 1.24 | 0.79 | 0.61 | 0.71 |

1) Commercial, 2) Water area, 3) Agriculture, 4) Green space and square, 5) Regional transport facilities, 6) Industrial, 7) Residential one, 8) Residential two, 9) Residential three, 10) Road, street and transportation, 11) Administration and public services.

Additionally, we also plotted the per pixel correlation between Wuhan in 2016 and Wuhan in 2014. As shown in Fig. 10, the mean absolute error of the derived and actual building density is 0.0063 and the correlation is 0.8708. The major area of Wuhan in 2016 and Wuhan in 2014 remains the same and the floor-area ratio mainly ranges between 0–3. The result confirms that using the deep learning method to learn from the massive data is efficient and has a reasonable accuracy.

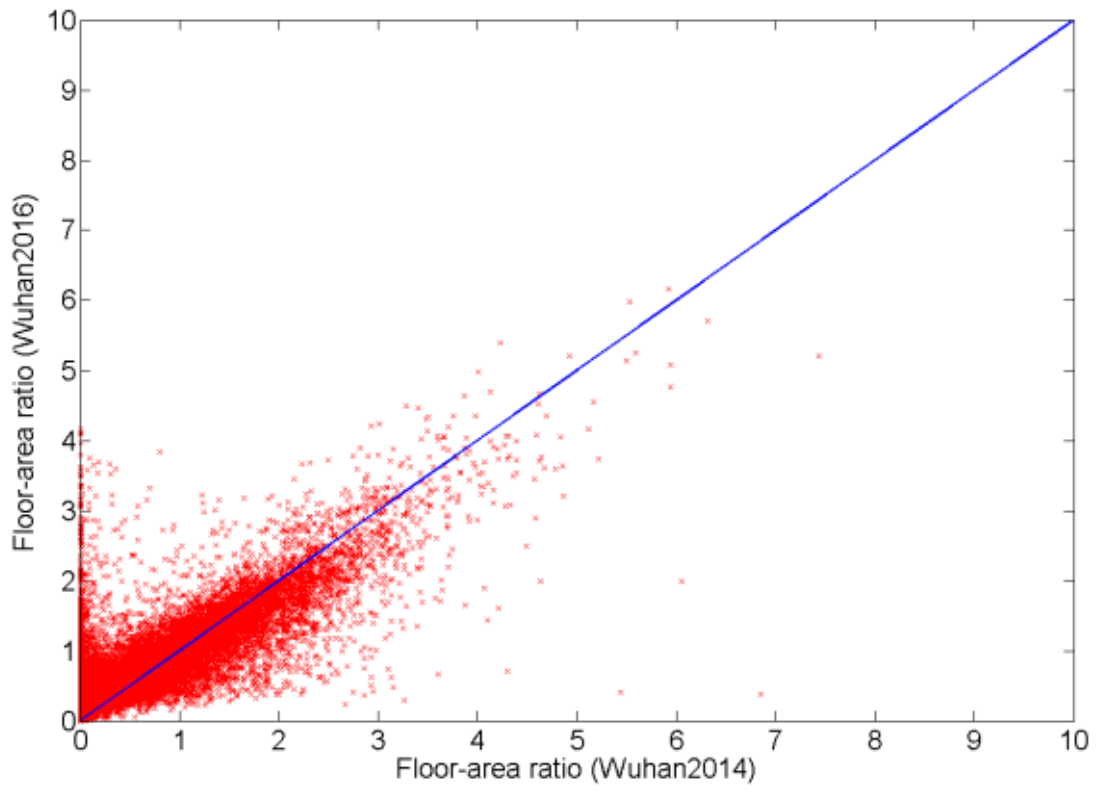

Fig. 10. Floor-area ratio correlation scatterplot between Wuhan 2016 and Wuhan 2014.

### *C. Population density estimation*

The accuracy of the derived population density map was assessed via the census data of Wuhan from 2014. However, the very high spatial resolution remote sensing data (acquired in January 2016) and census data (acquired in 2014) were obtained at different times, and the census data have their own margin of error. Therefore, the accuracy, in this study, mainly indicate the similarity to the census data.

The resulting map of the population density for the city of Wuhan is shown in Fig. 11. The per pixel correlation between Wuhan in 2016 and Wuhan in 2014 is shown in Fig. 12. The mean absolute error of the derived and actual population density is 209.7 and the correlation is 0.7212. The derived population density map can provide a more detailed population distribution for the whole of the city of Wuhan, even for the rural areas. For most of the central area of Wuhan, the proposed CNN model can reveal a similar pattern and population distribution. However, for the surrounding regions, the proposed model overestimates the population density since it only considers the land-use type and building structure and does not consider the occupancy rate. In fact, the distance to the central area significantly influences the occupancy rate. On the other hand, the census data in the surrounding regions are also difficult to obtain and may contain large errors. Therefore, the major discrepancy is concentrated in the surrounding regions.

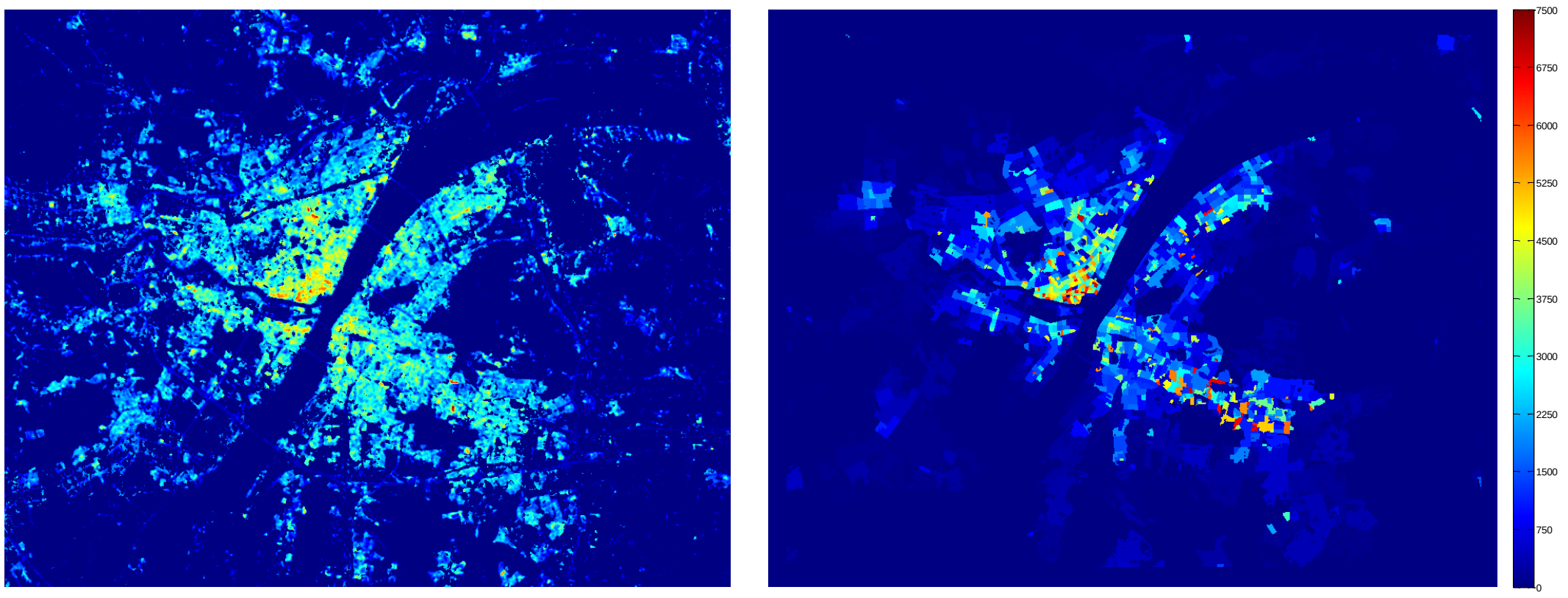

Fig. 11. Comparison of the population density. Left: the derived population density for the city of Wuhan in 2016. Right: the original census data for the city of Wuhan in 2014.

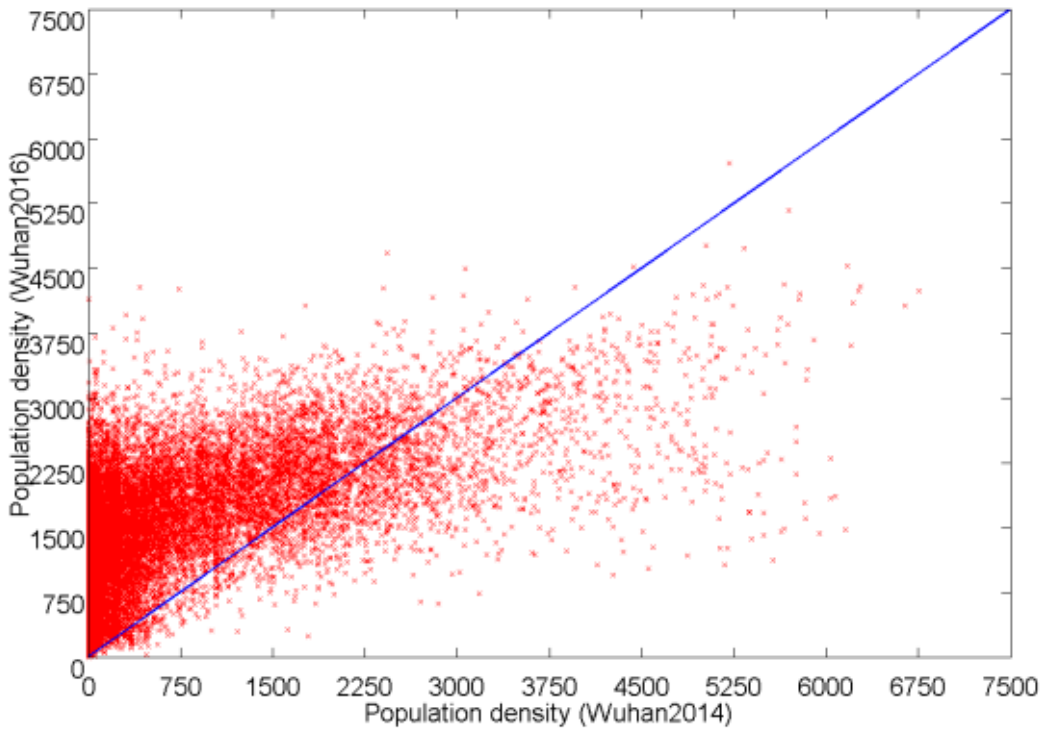

Fig. 12. Population density correlation scatterplot between Wuhan 2016 and Wuhan 2014.

## V. Discussion

### *A. Can we automatically learn the internal features from massive data for mega-city analysis?*

The study area covered a large urban region (2606 km$^2$, 47537 $\times$38100 pixels, 1.2 m spatial resolution) containing many natural topography, including water area, vegetation, and bare soil, as well as numerous artificial land use type, e.g., roads and many kinds of buildings (industrial buildings, commercial buildings, residential, stadiums, and village). Based on the related GIS data, we collected a large volume of training data for our model (26363 training images for land-use classification, 75733 training images for urban density estimation, and 98219 training images for population estimation). Compared to previous studies [6]-[9], this is the first time that such a large image has been tackled to perform a fine-scale analysis for a mega-city (land-use classification, urban density estimation, and population estimation).

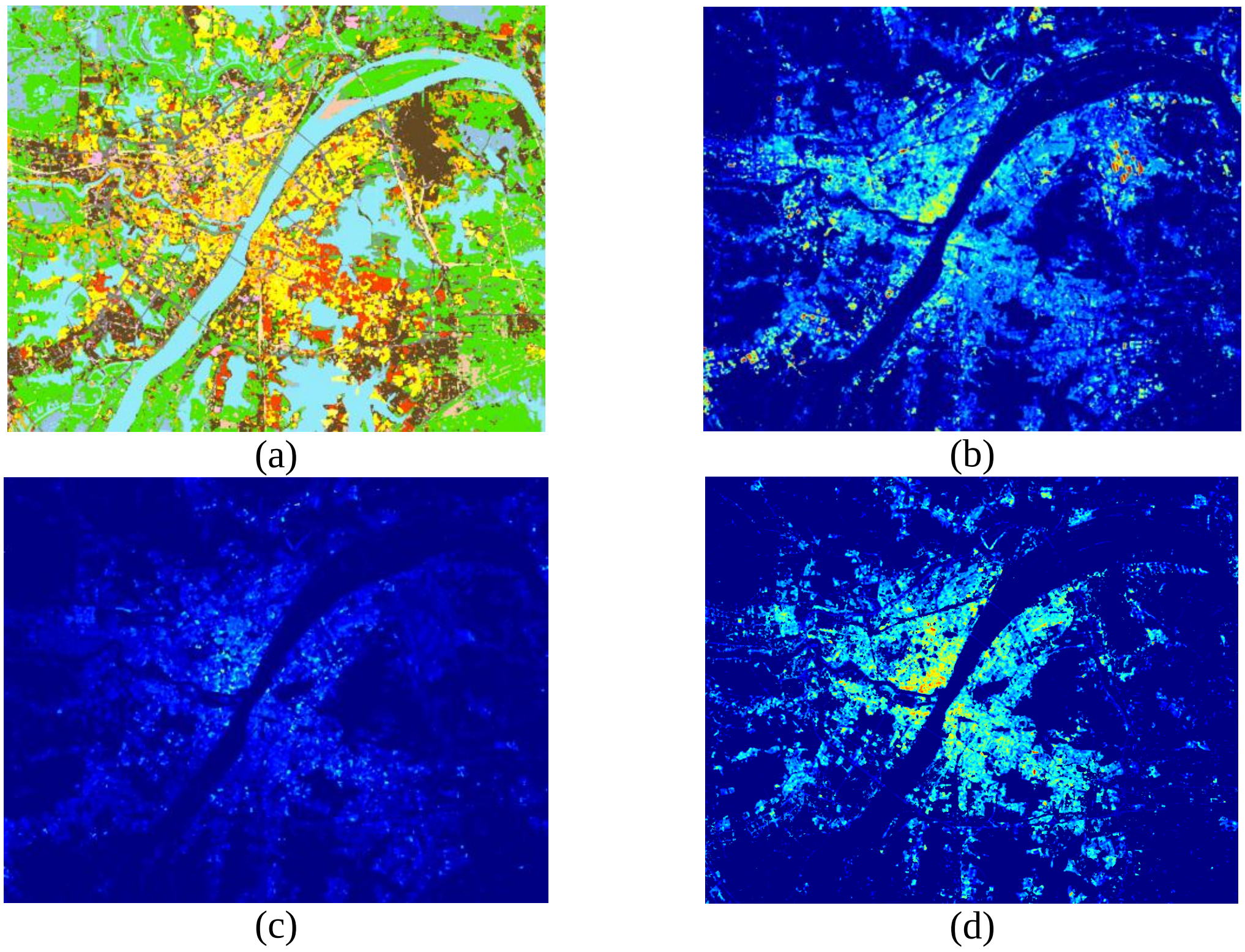

Fig. 13. The derived land-use map, building density map, floor-area ratio map, and population density map for the city of Wuhan in 2016: (a) land-use map; (b) building density map; (c) floor-area ratio map; (d) population density map.

From the large volume of training data, the proposed CNN model was able to automatically learn the internal feature representation and generate reasonable results. Combined with GPU acceleration, the training and processing time was also fast and efficient with consideration of the huge number of pixels (1,811,159,700 pixels). For the land-use classification task, the derived land-use map achieves a relatively high OA (92.35%) and reveals the major patterns for the different land-use types (Fig. 13). For the urban density estimation and population estimation, the mean absolute error and per pixel correlation also confirm that the proposed model can obtain reasonable results

Theoretically, using massive data to find the internal data structure and the feature representation for different tasks is efficient and accurate. Since the scenes are very complex and the feature learning stage is non-linear, the deep learning model can generate an accurate internal data structure. As a result, the hierarchical feature representation and deep learning method can automatically learn the end-to-end mapping from the massive data for mega-city analysis and can generate a reasonable degree of accuracy.

*B. The potential capacity of the CNN-based universal framework*

As reported in the experiments, the universal framework can generate promising land-use mapping results with regard to the OA and Kappa and can reveal the obvious patterns of urban density and population distribution. Furthermore, the universal framework breaks the barrier between different tasks and combines them into a single framework.

For remote sensing image analysis, there are several applications of the proposed model in fields such as land-use mapping, studies of the urban heat island effect, and population estimation. The previous works in remote sensing analysis have focused on a single task and have ignored the relationship between multiple tasks. Most of the traditional remote sensing image analysis methods consist of two stages: a feature extraction (calculation) stage and a classification stage. These two-stage methods ignore the relationship between the feature representation and the corresponding task. For every task, the previous studies have had to build different feature extraction (calculation) stages or feature selection stages to find the best feature representation.

In the proposed model, the universal framework combines the multiple tasks into a single framework, and the different tasks share the same feature learning stage, which dramatically reduces the human labor cost in the feature design. On the other hand, most remote sensing applications are highly related. The land-use type influences the urban density; the building density and floor-area ratio reflect the natural attributes of the land-use type; and population is highly correlated to socioeconomic factors. As shown in Table VI, for the different land-use types, the mean building density and mean floor-area ratio are different. The commercial type has the highest mean building density, which refers to the dense building structures and tall buildings Although the mean building density for “Residential three” is higher than “Residential two”, due to the poor living conditions and construction, “Residential three” has a lower mean floor-area ratio. The universal framework can further process the information between multiple tasks and provide an accurate result.

*C. The mega-city analysis of central Wuhan using a VHR remote sensing image*

In this study, we used a VHR remote sensing image to generate a land-use map, urban density map (building density and floor-area ratio), and population density map for the mega-city analysis. It is well known that different cities usually exhibit individual characteristics and unique features, although the structural similarities is still exist. For better city planning and management, it is useful and efficient to be able to produce such fine-scale land-use, urban density (building density and floor-area ratio), and population density maps for city monitoring and analysis.

Wuhan is the financial, political, educational, and transportation center of central China. The derived land-use map reveals the major patterns and the land-use spatial distribution of Wuhan, such as the road network and water area distribution.

Wuhan is positioned at the confluence of the Yangtze River and Han River, and is delineated by the two rivers into three towns, i.e., Wuchang, Hankou, and Hanyang (Fig. 1). Most of the commercial areas are located in Hankou, and the educational areas, such as the university campus, are mainly found in Wuchang (Fig. 13a). Due to the dense commercial area, the building density of Hankou is higher than that of Wuchang (Fig. 13b–c). The better commercial development conditions also attract more people to live in Hankou, which results in a higher population density than the other areas (Fig. 13d). However, because of the rapid development and urbanization, a significant proportion of Hankou is classified as "Residential three". In summary, such fine-scale land-use, urban density (building density and floor-area ratio), and population density maps can provide us with a better understanding of the mega-city structure.

## VI. Conclusion

In this study, we have applied a deep learning method to automatically learn the internal feature representation from massive data for a mega-city analysis. The land-use classification, urban density estimation, and population estimation provide the basis for effective city monitoring and management. A CNN-based universal framework for multiple tasks has been proposed, enabling the holistic analysis of a mega-city. The use of the VHR remote sensing image and deep learning method offer objectivity, efficiency, and automation with regard to the large city area. Compared to previous works which have needed elaborate manually designed frameworks and comprehensive feature calculation, this is the first time that a CNN-based universal framework has been used to automatically learn the features from massive data and combine multiple tasks into a single framework. Satisfactory land-use classification results and urban density and population maps were simultaneously produced by the proposed method. Based on the change analysis of the Wuhan 2014 and Wuhan 2016 VHR images, distinct differences are apparent, which have mainly been caused by the rapid urbanization and the change in land-use types. The per pixel correlation and the mean absolute error of the Wuhan 2014 and Wuhan 2016 VHR images also reveal the same result, in that the spatial change is mainly located in the surrounding areas of Wuhan.

## Acknowledgments

The authors would like to specifically thank Jiong Wang from the School of Urban Design, Wuhan University, for providing the land-use data, urban density data and census data.

## REFERENCES

[1] F. Kraas, "Mega cities and global change: Key priorities," *Geogr. J.,* vol. 173, no. 1, pp. 79–82, 2007.

[2] R. Fuchs, E. Brennan, J. Chamie and J. Uitto, "Megacity growth and the future," United Nations University Press, Tokyo., 1994.

[3] W. Li, C. Chen, H. Su, et al, "Local binary patterns and extreme learning machine for hyperspectral imagery classification," *IEEE Trans. Geosci. Remote Sens*., vol. 53, no. 7, pp. 3681–3693, 2015.

[4] J. Mitchell, "Mega cities and natural disasters: A comparative analysis," *Geo J.,* vol. 49, pp. 137-142, 1999.

[5] S. Pauleit and F. Duhme, "Assessing the environmental performance of land cover types for urban planning," *Landscape Urban Plann.,* vol. 52, no. 1, pp. 1–20, 2000.

[6] X. Zhang and S. Du, "A linear dirichlet mixture model for decomposing scenes: Application to analyzing urban functional zonings," *Rem. Sens. Environ.*, vol. 169, pp. 37-49, 2015.

[7] H. Taubenböck, M. Klotz, M. Wurm, *et al*., "Delineation of central business districts in mega city regions using remotely sensed data," *Rem. Sens. Environ.*, vol. 136, pp. 386-401, 2013.

[8] J. Susaki, M. Kajimoto and M. Kishimoto, "Urban density mapping of global megacities from polarimetric SAR images," *Rem. Sens. Environ.*, vol. 155, pp. 334-348, 2014.

[9] D. Azar, R. Engstrom, J. Graesser and J. Comenetz, "Generation of fine-scale population layers using multi-resolution satellite imagery and geospatial data," *Rem. Sens. Environ.*, vol. 130, pp. 219-232, 2013.

[10] F. Pacifici, M. Chini and W. J. Emery, "A neural network approach using multi-scale textural metrics from very high-resolution panchromatic imagery for urban land-use classification," *Rem. Sens. Environ.*, vol. 113, pp. 1276-1292, 2009.

[11] A. Schneider, "Monitoring land cover change in urban and peri-urban areas using dense time stacks of Landsat satellite data and a data mining approach," *Rem. Sens. Environ.*, vol. 124, pp. 689–704, 2012.

[12] Y. Tian, T. Yue, L. Zhu and N. Clinton, “Modeling population density using land cover data,” *Ecol. Model.*, vol. 189, no. 11, pp. 72-88, 2005.

[13] J. Wang, Z. Qingming, H. Guo and Z. Jin, “Characterizing the spatial dynamics of land surface temperature–impervious surface fraction relationship,” *Int. J. Appl. Earth Obs. Geoinf.*, vol. 45, pp. 55-65, 2016.

[14] Y. Yang and S. Newsam, “Bag-of-visual-words and spatial extensions for land-use classification,” in *Proc. ACM Int. Conf. Adv. Geogr. Inf. Syst.*, pp. 270-279, 2010.

[15] F. Hu, G. Xia, Z. Wang, X. Huang, L. Zhang, and H. Sun, “Unsupervised Feature Learning Via Spectral Clustering of Multidimensional Patches for Remotely Sensed Scene Classification,” *IEEE J. Sel. Topics Appl. Earth Observ. Remote Sens*., vol. 8, no. 5, pp. 2015–2030, May 2015.

[16] M. Voltersen, C. Berger, S. Hese and C. Schmullius, “Object-based land cover mapping and comprehensive feature calculation for an automated derivation of urban structure types at block level,” *Rem. Sens. Environ.*, vol. 154, pp. 192-201, 2014.

[17] L. Zhang, L. Zhang and B. Du, “Deep Learning for Remote Sensing Data: A Technical Tutorial on the State of the Art,” *IEEE Geosci. Remote Sens. Mag.*, vol. 4, no. 2, pp. 22-40, 2016.

[18] A. Schneider, M. A. Friedl and D. Potere, “Mapping global urban areas using MODIS 500-m data: New methods and datasets based on ‘urban ecoregions,” *Rem. Sens. Environ.*, vol. 114, pp. 1733–1746, 2010.

[19] X. Niu and Y. Ban, “An adaptive contextual SEM algorithm for urban land cover mapping using multitemporal high-resolution polarimetric SAR data,” *IEEE J. Sel. Topics Appl. Earth Observ. Remote Sens*., vol. 5, pp. 1129–1139, 2012.

[20] M. Kajimoto and J. Susaki, “Urban density estimation from polarimetric SAR images based on a POA correction method,” *IEEE J. Sel. Topics Appl. Earth Observ. Remote Sens*., vol. 6, pp. 1418–1429, 2013

[21] G. Hinton, L. Deng, D. Yu, *et al.*, “Deep neural networks for acoustic modeling in speech recognition: The shared views of four research groups,” *IEEE Signal Process. Mag.*, vol. 29, pp. 6, pp. 82-97, 2012.

[22] C. Farabet, C. Couprie, L. Najman and Y. LeCun, “Learning hierarchical features for scene labeling,” *IEEE Trans. Pattern Anal. Mach. Intell.*, vol. 35, no. 8, pp. 1915-1929, 2013.

[23] P. Sermanet, D. Eigen, X. Zhang, M. Mathieu, R. Fergus, and Y. LeCun. “OverFeat: Integrated Recognition, Localization and Detection using Convolutional Networks,” in *Proc. ICLR*, 2014.

[24] D. Eigen, C. Puhrsch and R. Fergus, “Depth map prediction from a single image using a multi-scale deep network,” *Adv. Neur. In*., pp. 2366-2374, 2014.

[25] A. Krizhevsky, I. Sutskever, and G. Hinton, “ImageNet classification with deep convolutional neural networks,” *Adv. Neur. In*., vol. 25, pp. 1106–1114, 2012.

[26] F. Zhang, B. Du, and L. Zhang, “Scene Classification via a Gradient Boosting Random Convolutional Network Framework,” *IEEE Trans. Geosci. Remote Sens*, vol. 54, no. 3, pp. 1793–1802, 2016.

[27] J. Yuan, “Automatic Building Extraction in Aerial Scenes Using Convolutional Networks,” arXiv preprint arXiv:1602.06564, 2016.

[28] Ministry of Housing and Urban-Rural Development of the People’s Republic of China (MOHURD), “Code for Classification of Urban Land Use and Planning Standards of Development Land,” China Architecture & Building Press, 2012.

[29] H. Chung, “Building an image of Villages‐in‐the‐City: A Clarification of China's Distinct Urban Spaces,” *Int. J. Urban Regional Res.*, vol. 34, no. 2, pp. 421-437, 2010.

[30] Y. LeCun, B. Boser, J. S. Denker, D. Henderson, R. E. Howard, W. Hubbard, and L. D. Jackel, “Handwritten digit recognition with a back-propagation network,” *Adv. Neur. In*., 1990.

[31] Y. LeCun, L. Bottou, Y. Bengio, and P. Haffner, “Gradient-based learning applied to document recognition,” *Proc. IEEE*, 1998.

[32] S. Ren, K. He, R. Girshick, *et al*., “Faster R-CNN: Towards real-time object detection with region proposal networks,” *Adv. Neur. In*, pp. 91–99, 2015.

[33] Y. Wang, L. Zhang, X. Tong, *et al.*, “A Three-Layered Graph-Based Learning Approach for Remote Sensing Image Retrieval,” *IEEE Trans. Geosci. Remote Sens.*, vol. 54, no. 10, pp. 6020 – 6034, 2016.

[34] T. Chen, M. Li, Y. Li, *et al*., “Mxnet: A flexible and efficient machine learning library for heterogeneous distributed systems,” arXiv preprint arXiv:1512.01274, 2015.